\documentclass[letterpaper, 10 pt, journal, twoside]{IEEEtran} 
\IEEEoverridecommandlockouts

\usepackage{times}
\usepackage[pdftex]{graphicx}
\usepackage{graphics}
\usepackage[cmex10]{amsmath}
\usepackage{url}
\usepackage{array}
\usepackage{tabularx}
\usepackage{multirow}
\usepackage{multicol}
\usepackage{booktabs}
\usepackage{epstopdf}
\usepackage{rotating}
\usepackage{csquotes}

\usepackage[export]{adjustbox}
\usepackage{amssymb}
\usepackage{amsmath}
\usepackage[all]{xy}    
\usepackage{ifthen}
\usepackage[usenames,dvipsnames]{color}
\usepackage{colortbl}
\usepackage{enumerate}
\usepackage{bm}
\usepackage{algorithm}
\usepackage{algorithmic}

\makeatletter
\let\NAT@parse\undefined
\makeatother
\usepackage[numbers]{natbib}

\newcommand{\eref}[1]{(\ref{#1})}
\newcommand{\secref}[1]{Section~\ref{#1}}
\newcommand{\tabref}[1]{Table~\ref{#1}}

\newcommand{\figref}[1]{Fig.~\ref{#1}}

\newcommand{\myparagraph}[1]{\vspace{0.1in}\noindent\textbf{#1}}

\newboolean{draft-mode}
\setboolean{draft-mode}{false}
\newcommand{\sidenote}[1]{\ifthenelse{\boolean{draft-mode}}{\marginpar{\tiny\raggedright\textsf{\hspace{0pt}#1}}}{}}
\DeclareRobustCommand{\arnote}[1]{\ifthenelse{\boolean{draft-mode}}{\textcolor{blue}{\textbf{AR: #1}}}{}}
\DeclareRobustCommand{\dmnote}[1]{\ifthenelse{\boolean{draft-mode}}{\textcolor{cyan}{\textbf{DM: #1}}}{}}

\newboolean{review-mode}
\setboolean{review-mode}{false}
\DeclareRobustCommand{\change}[1]{\ifthenelse{\boolean{review-mode}}{\textcolor{red}{#1}}{ \textcolor{black}{#1} }}

\title{Friction Variability in Planar Pushing Data:\\  Anisotropic Friction and Data-collection Bias
}

\author{Daolin Ma$^{1}$, and Alberto Rodriguez$^{1}$%
\thanks{Manuscript received: February 24, 2018; Revised May, 17, 2018; Accepted June, 14, 2018.}
\thanks{This paper was recommended for publication by Editor Han Ding upon evaluation of the Associate Editor and Reviewers' comments. 
This work was supported by NSF award [IIS-1637753] through the National Robotics Initiative. Daolin Ma is partially supported by the National Natural Science Foundation of China (NSFC) under grant No.51608458 and a scholarship from the China Scholarship Council (CSC).} 
\thanks{$^{1}$Daolin Ma and Alberto Rodriguez are with Mechanical Engineering Department --- Massachusetts Institute of Technology
         {\tt\footnotesize <daolinma,albertor>@mit.edu}}%
\thanks{Digital Object Identifier (DOI): see top of this page.}
}

\begin{document}
\maketitle

\begin{abstract}
Friction plays a key role in manipulating objects.
Most of what we do with our hands, and most of what robots do with their grippers, is based on the ability to control frictional forces.
This paper aims to better understand the variability and predictability of planar friction.
In particular, we focus on the analysis of a recent dataset on planar pushing by \citet{Yu2016MorePushing} devised to create a data-driven footprint of planar friction. 

We show in this paper how we can explain a significant fraction of the observed unconventional phenomena, e.g., stochasticity and multi-modality, by combining the effects of material non-homogeneity, anisotropy of friction and biases due to data collection dynamics, hinting that the variability is explainable but inevitable in practice.

We introduce an anisotropic friction model and conduct simulation experiments comparing with more standard isotropic friction models. The anisotropic friction between object and supporting surface results in convergence of initial condition during the automated data collection. Numerical results confirm that the anisotropic friction model explains the bias in the dataset and the apparent stochasticity in the outcome of a push.
%
%
The fact that the data collection process itself can originate biases in the collected datasets, resulting in deterioration of trained models, calls attention to the data collection dynamics. 

\end{abstract}

\section{Introduction}

\IEEEPARstart{U}{nderstanding} friction between two sliding surfaces has been a focus of research for many decades. Even under carefully controlled experimental conditions, planar friction manifests a surprising degree of variability, which makes it difficult to model and control. 
This paper studies the nature of that variability in a robotics context. In particular we focus on the analysis of a recent dataset of sliding frictional interactions by~\citet{Yu2016MorePushing} and~\citet{Bauza2017APushing}. Our goal is to explain the nature of the observed variability (e.g., see \figref{fig: Setup and Structured uncertainty}) and suggest models to account for it.

Our analysis shows that a detailed understanding of the experimental process used to collect the data is key. Indeed data-driven modeling and automatic data collection are widespread techniques in robotics. 
%
%
In contact tasks, which involve complex and weakly observed dynamics, it is difficult to avoid data variability due to sensor noise, experiment bias, and task variations. Understanding the nature of that variability is an important step to build reliable data-driven models. 

\secref{sec:facts} describes experimental facts observed in Yu \textit{et al.}'s dataset \cite{Yu2016MorePushing}, including the dependence of the data dynamics with the initial orientation of the pushed object, the pronounced \emph{variability} of the pushes, and the bias observed in the dataset.
In particular we focus on sequences of 2000 repeated pushes, part of the dataset, that lead to outcomes with significant variability, as in \figref{fig: Setup and Structured uncertainty}. These point to three key questions that we attempt to answer in this paper:

\begin{figure}[t]
    \centering
    \includegraphics[width=\linewidth]{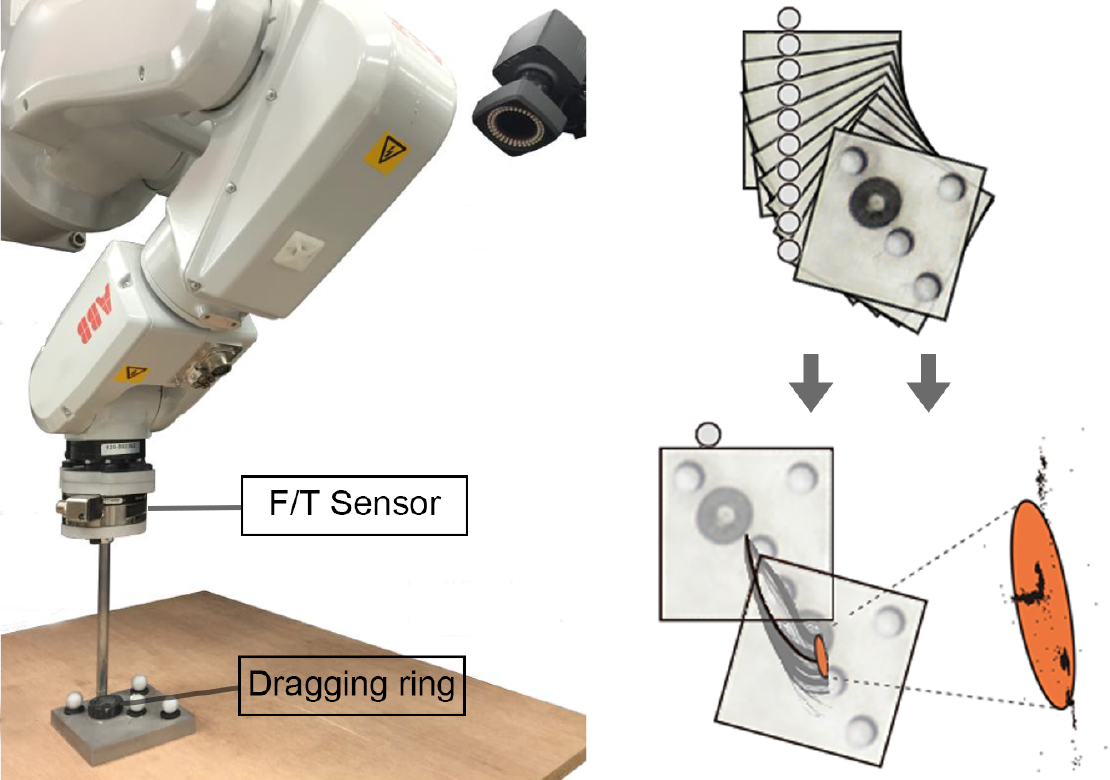}
    \caption{\change{(\textbf{Left}) Experiment setup of repeated pushing from~\cite{Yu2016MorePushing}. The robot arm holds a cylinder that pushes a rectangular object along a straight line. 
    (\textbf{Right}) Trajectories of the center of mass of the object during 2000 pushes. The figure, adapted from~\citep{Bauza2017APushing}, shows an interesting distribution of end poses.}}
    \label{fig: Setup and Structured uncertainty}
\end{figure}

\subsection{Why is pushing not behaving determiniscally?}
Pushing, a task driven by the dynamics of planar friction, is a key manipulation primitive that humans and robots exploit to manipulate objects.
The robotics community has developed a series of models to describe its dynamics~\cite{Mason1986MechanicsOperations,Peshkin1988MotionWorkpiece,Lynch1992ManipulationFeedback,Jia1999PoseContact,Liu2011PushingModel} relying on a deterministic Coulomb friction law, naturally resulting in a determined pushed motion. These models have been widely used to develop control strategies~\cite{Lynch1992ManipulationFeedback,Lynch1996StablePlanning,Behrens2013RoboticTechniques,Hogan2016FeedbackDynamics}. 
The accuracy of these models remains in question since the modelling and mechanism of friction is still an open problem~\cite{Marone1998TheCycle, Mate1987,Bylinskii2015,Yamashita2015}.

If it is natural to explain friction as a deterministic, albeit weakly observed, process, how do we explain the significant and structured variability of trajectories in \figref{fig: Setup and Structured uncertainty}?
Although it is possible---and maybe practical for robotics---to describe that variability lumped into a stochastic process~\cite{Bauza2017APushing,Zhou2017AValidation}, in this paper we show that it is also possible to explain most of the variability in simple mechanical terms.
\secref{sec:formation_uncertainty} describes the formation of the path variability for four different materials: \emph{plywood}, \emph{abs}, \emph{delrin}, and \emph{polyurethane}.
%

\subsection{What is the origin of the structure in the noise?}


\citet{Yu2016MorePushing} describes the variability of friction in the dataset with respect to changes in the location, direction, time, and speed of a push. These account for a fraction of the necessary discrepancy against simple models of friction.
Note however that \figref{fig: Setup and Structured uncertainty} shows a marked non-Gaussian complex structure, which is more difficult to explain with mechanical arguments of simple variations of the coefficient of friction. Does the effective friction coefficient follow a complex distribution?

In this paper, we show that it is not necessary to invoke stochastic explanations, and that the multi-modality structure can be explained by a combination of anisotropic friction and bias in the data.
\secref{sec:formation_bias} describes the process by which the bias in the dataset is formed by looking at the data collection as a dynamic process. This leads to stable and unstable pushing directions that affect the distribution of initial orientations of the pushed object.

\subsection{How can we model friction more accurately?}


In recent years, we have seen a significant number of data-driven models of friction and pushing~\citep{Mericli2014Push-manipulationModels,Salganicoff1993MarcosSandini,Lau2011AutomaticObjects,Walker2008PushingMaps,Zhou2016AApplication, Bauza2017APushing} as an alternative to analytic ones.
\citet{Zhou2017AValidation} assume that the contact behavior follows a generalized Coulomb friction law, but that the coefficient of friction is a random variable following a Gaussian distribution. Numerical results show that this friction model can simulate some of the variability. Following the observation that different pushes yield different levels of uncertainty, \citet{Bauza2017APushing} propose a purely data-driven probabilistic model based on Heteroscedastic Gaussian Processes that predicts the mean and expected variance of the motion of a pushed object. 

There are simpler mechanical explanations which require to account for anisotropic friction and biases in the dataset. 
In this paper, we propose an anisotropic friction model and simulate the data collection dynamics with it. The simulation can reproduce the bias and the direction convergence that is observed in experiment data and confirms that anisotropic friction is a key source of variability that can explain the multi-modality.

\section{Review of experimental setup}
\label{sec:Setup}
Before starting the analysis, we review the experiment setup used to collect the data.




\subsection{Experiment setup}
\label{sec:experiment_setup}
To enable the collection of a large data-set, \citet{Yu2016MorePushing} automated a loop of \textit{pushing} \& \textit{re-positioning} an object, with pushes carefully executed in the initial reference frame of the object. 
A subset of the experiments yielded continuous runs of 2000 identical pushes over four different surfaces made of \textit{plywood, abs, delrin} and \textit{pu}. Each experiment is composed of two phases:


\begin{itemize}
    \item \textbf{Pushing phase}: The robotic arm fitted with a thin cylindrical rod (see \figref{fig: Setup and Structured uncertainty}), pushes the object along a predetermined trajectory relative to the initial pose of the object.
    In the set of experiments we analyze in the paper, the pushing direction starts orthogonal to the object edge, as illustrated in \figref{fig: Setup and Structured uncertainty}. In the particular experiment that leads to the distribution, the pushing distance is 150mm with velocity 20mm/s. 

    \item \textbf{Re-positioning phase}: After a pushing phase, the robot drags the object back to the central area of the surface by inserting the thin rod in a ring attached to the object and dragging it. The dragging ring is located off-center, as shown in \figref{fig: Setup and Structured uncertainty}.
\end{itemize}

\begin{figure}[t]
    \centering
    \includegraphics[width=0.95\linewidth]{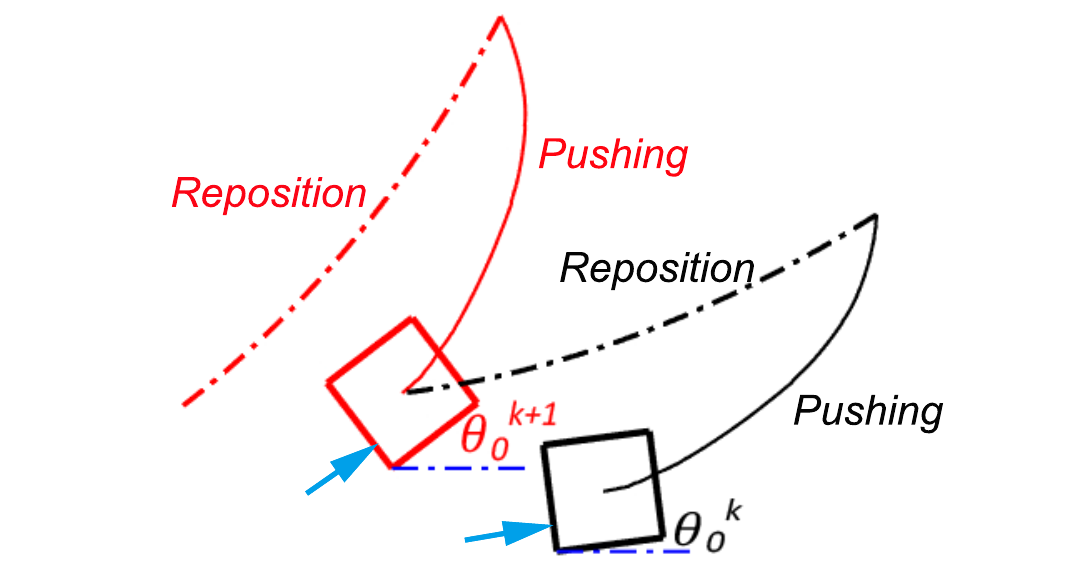}
    \caption{Illustration of data collection dynamics. The path shows the motion of the CoM of the block during the pushing (solid) and re-positioning (dashed) phases for two consecutive loops of the experiment $k$ and $k+1$. The arrows show the initial pushing locations and directions $\theta_0^k$ and $\theta_0^{k+1}$, which are identical in the initial frame of the object but vary in the global reference frame.} 
    \label{fig:Illustration_of_data_collection_dynamics}
\end{figure}
\subsection{Structured variability and notation}

As illustrated in \figref{fig:Illustration_of_data_collection_dynamics}, the initial position and orientation of the object at the beginning of each push varies. The re-positioning phase brings the object close but, not exactly, to the center of the surface.
This is done for simplicity in automating the experiments, but also because it naturally generates a richer dataset.

For analyzing the results, we denote Initial Object Frame (IOF) as the initial frame of the object before it is pushed. In that frame, the object starts at $(x, y, \theta) = (0, 0, 0)$ and ends at $(x, y, \theta) = (\Delta x, \Delta y, \Delta \theta)$.
%

\begin{figure}[t]
    \begin{minipage}[t]{0.50\linewidth}
    \centering
    plywood
    \includegraphics[width=\linewidth]{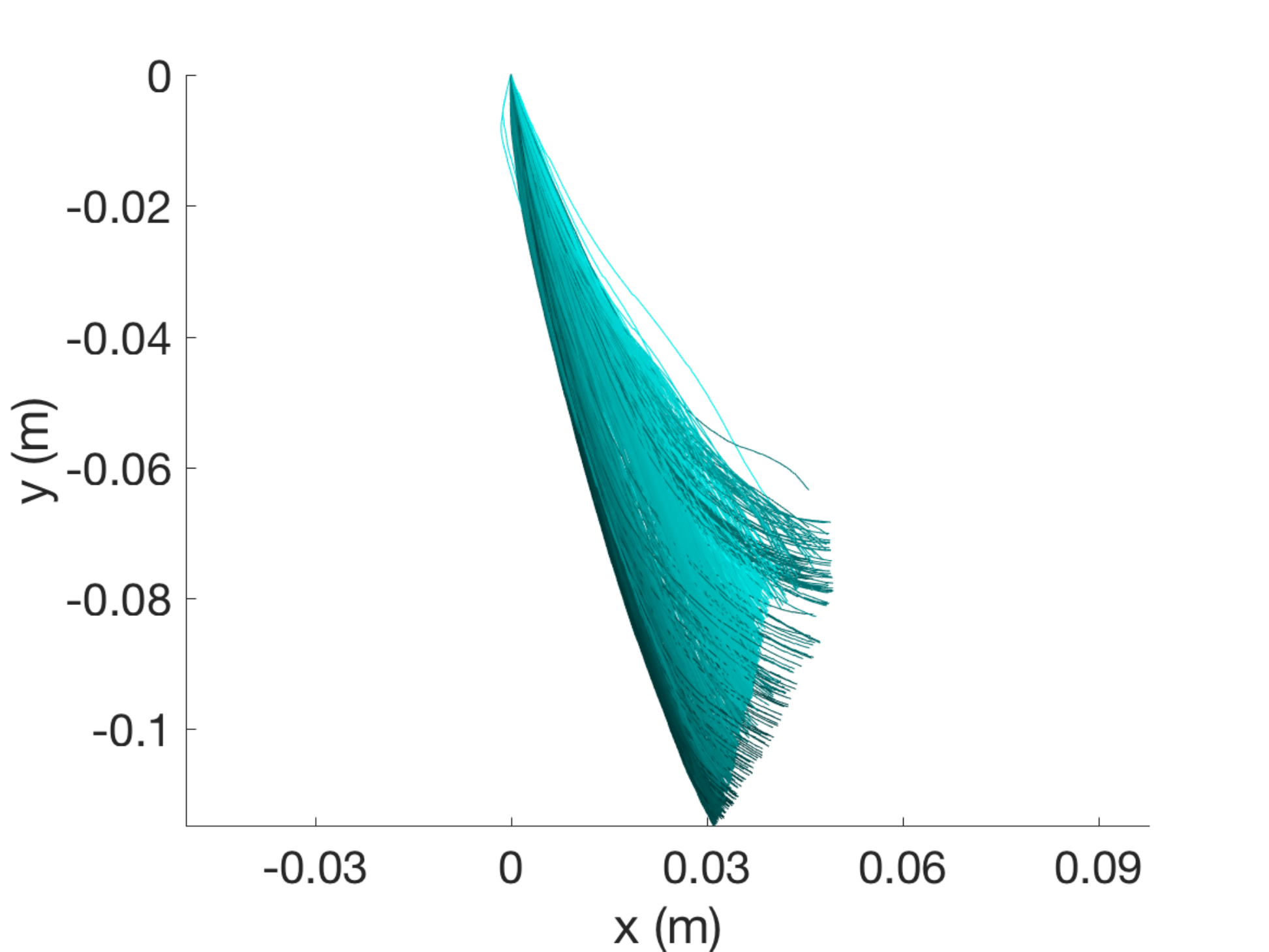}
    \end{minipage}%
    \begin{minipage}[t]{0.50\linewidth}
    \centering
    abs
    \includegraphics[width=\linewidth]{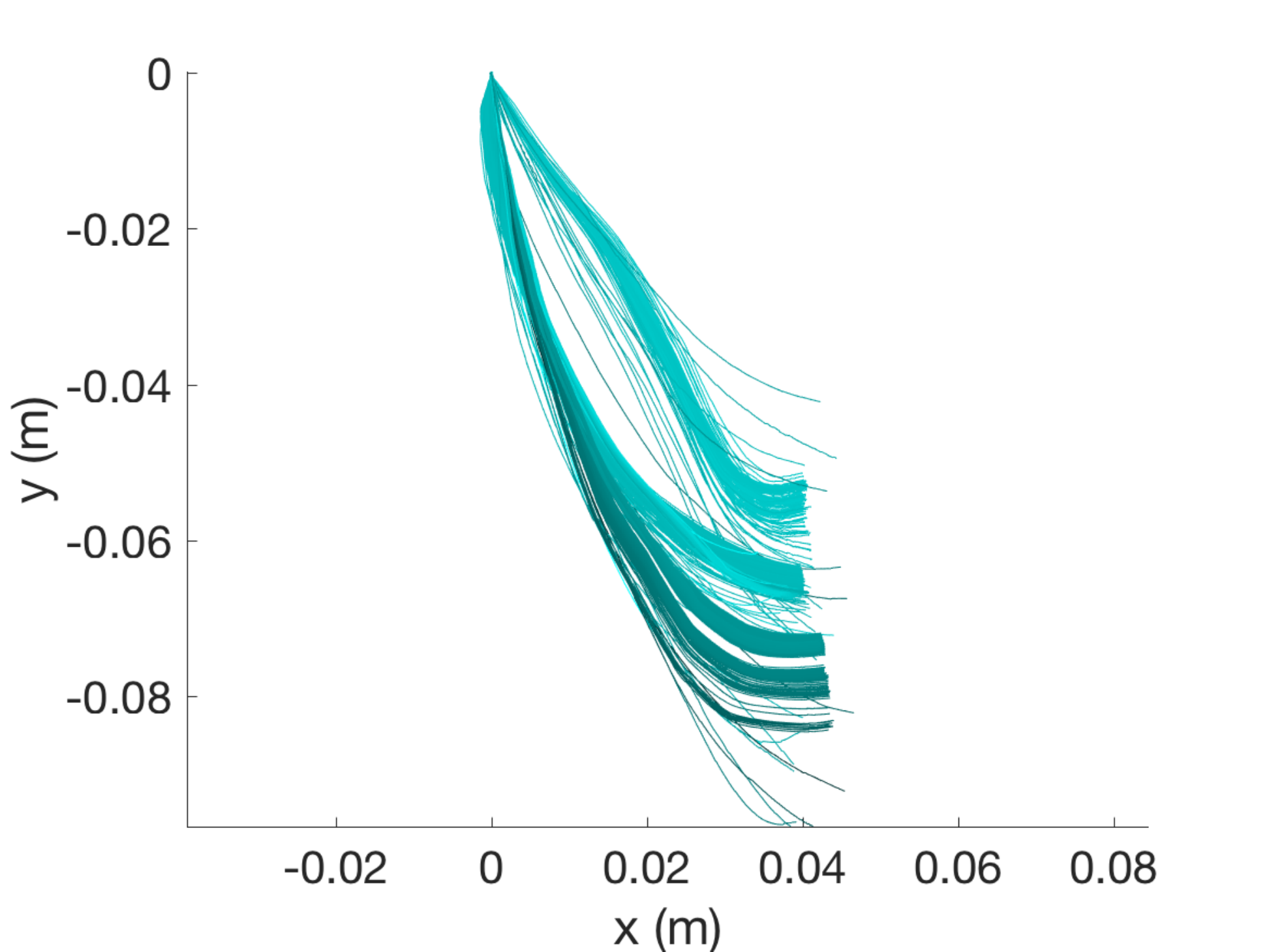}
    \end{minipage}
    \vspace{.2cm} 
    \begin{minipage}[t]{0.49\linewidth}
    \centering
    delrin
    \includegraphics[width=\linewidth]{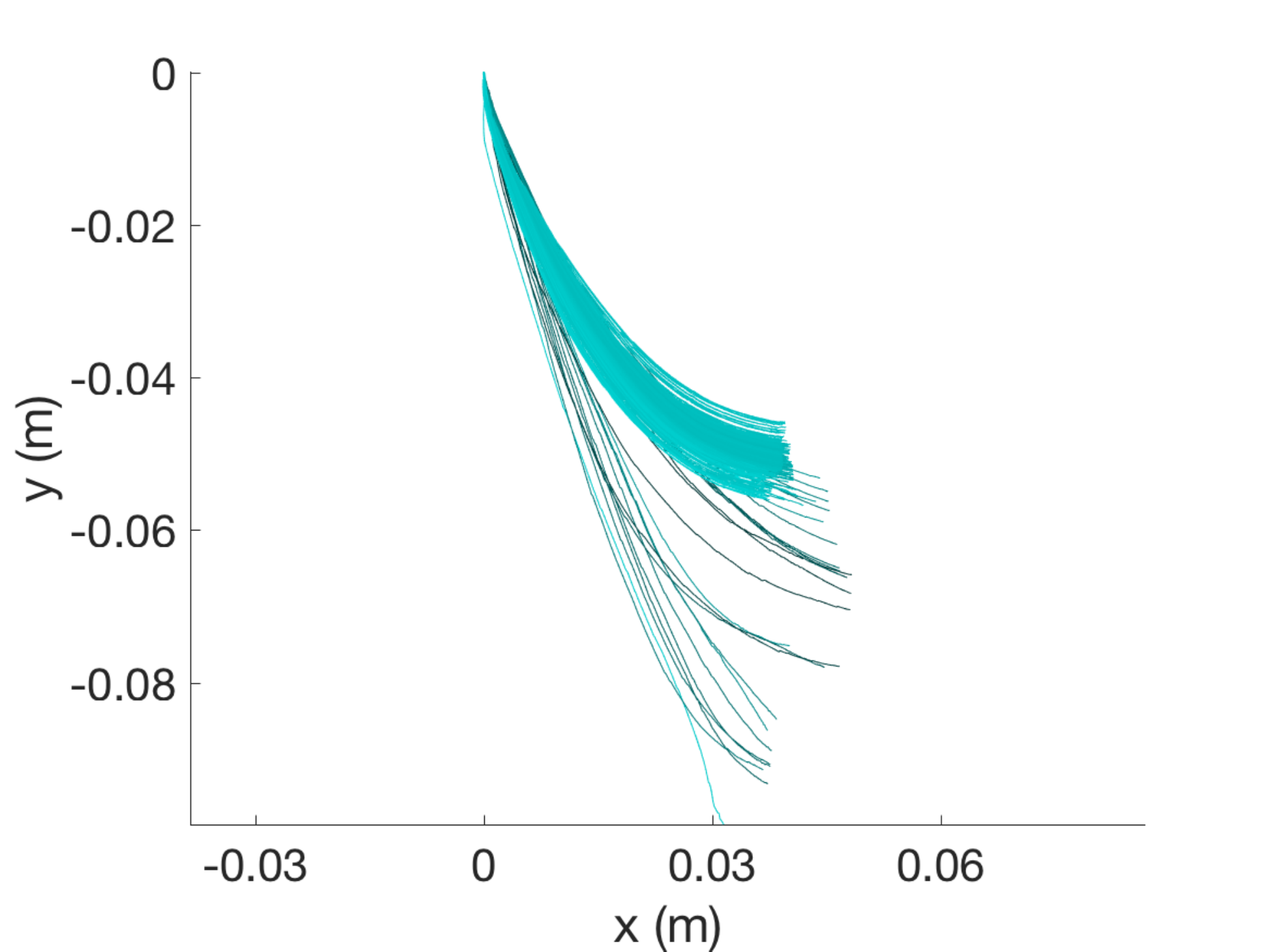}
    \end{minipage}
    \begin{minipage}[t]{0.49\linewidth}
    \centering
    pu
    \includegraphics[width=\linewidth]{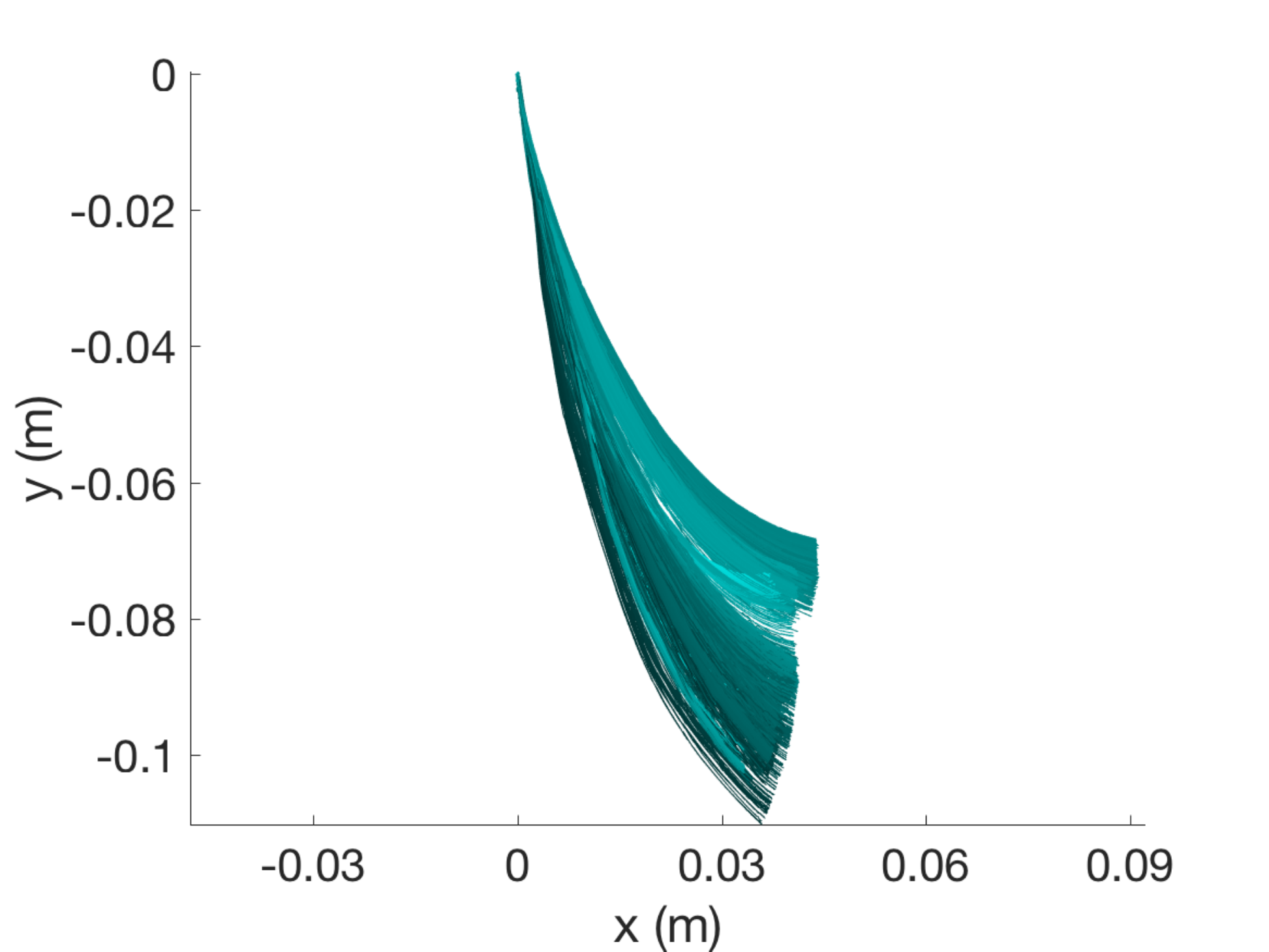}
    \end{minipage}
    \caption{Trajectories of the CoM of the pushed object in its reference frame on four different materials: plywood, abs, delrin, and polyurethane. A darker shade represents larger change in $\Delta\theta$. Note the complex---sometimes multimodal---structure of the distribution of paths.}
    \label{fig:Trajectories of CM in Object Base}
\end{figure}

Figures~\ref{fig:Trajectories of CM in Object Base} and \ref{fig:Trajectories of CM in Robot Base} illustrate the paths followed by the CoM of the object during 2000 pushes on 4 different materials, plotted both in IOF and in the global reference frame respectively.
The structured variability in experiment becomes clear in Fig.~\ref{fig:Trajectories of CM in Object Base}. 

\section{Analysis of experimental data}
\label{sec:facts}
In this section we first describe facts observed in the dataset and then show how they play a key role in the formation of structured variability. Finally, we analyze the data collection dynamics to explain biases in initial conditions.

\subsection{Facts}
\myparagraph{Dependence on initial orientation.}
Simulators frequently assume that friction is homogeneous and isotropic. In our particular scenario, that would mean that the trajectory of a pushed object in the initial reference frame of a push (IOF) is independent of its initial position and orientation.
This turns out to be far from reality.

\begin{figure}[t]
    \begin{minipage}[t]{0.5\linewidth}
   
    \centering
    {plywood}
    \includegraphics[width=\linewidth]{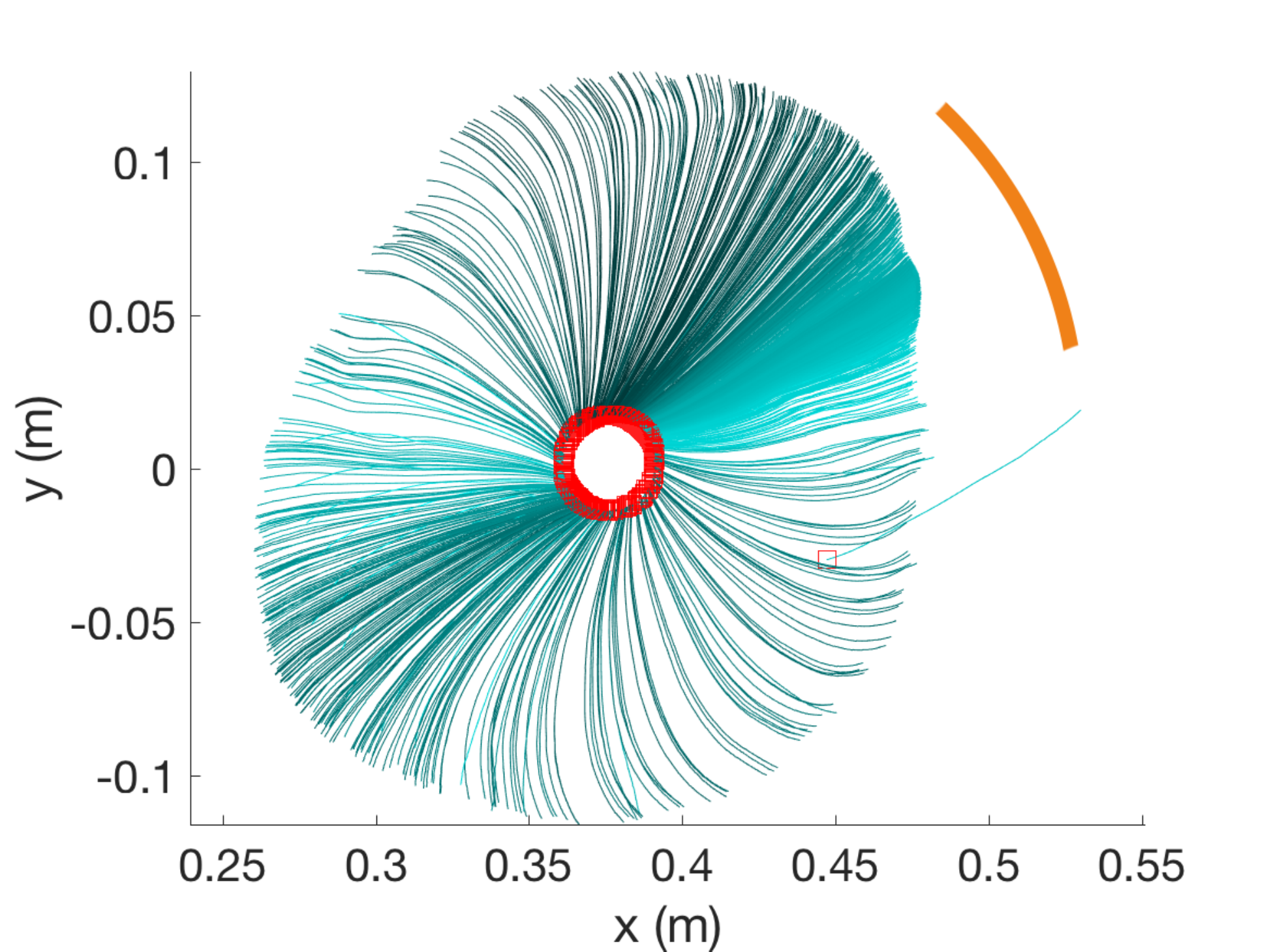}
    \end{minipage}%
    \begin{minipage}[t]{0.5\linewidth}
    \centering
    {abs}
    \includegraphics[width=\linewidth]{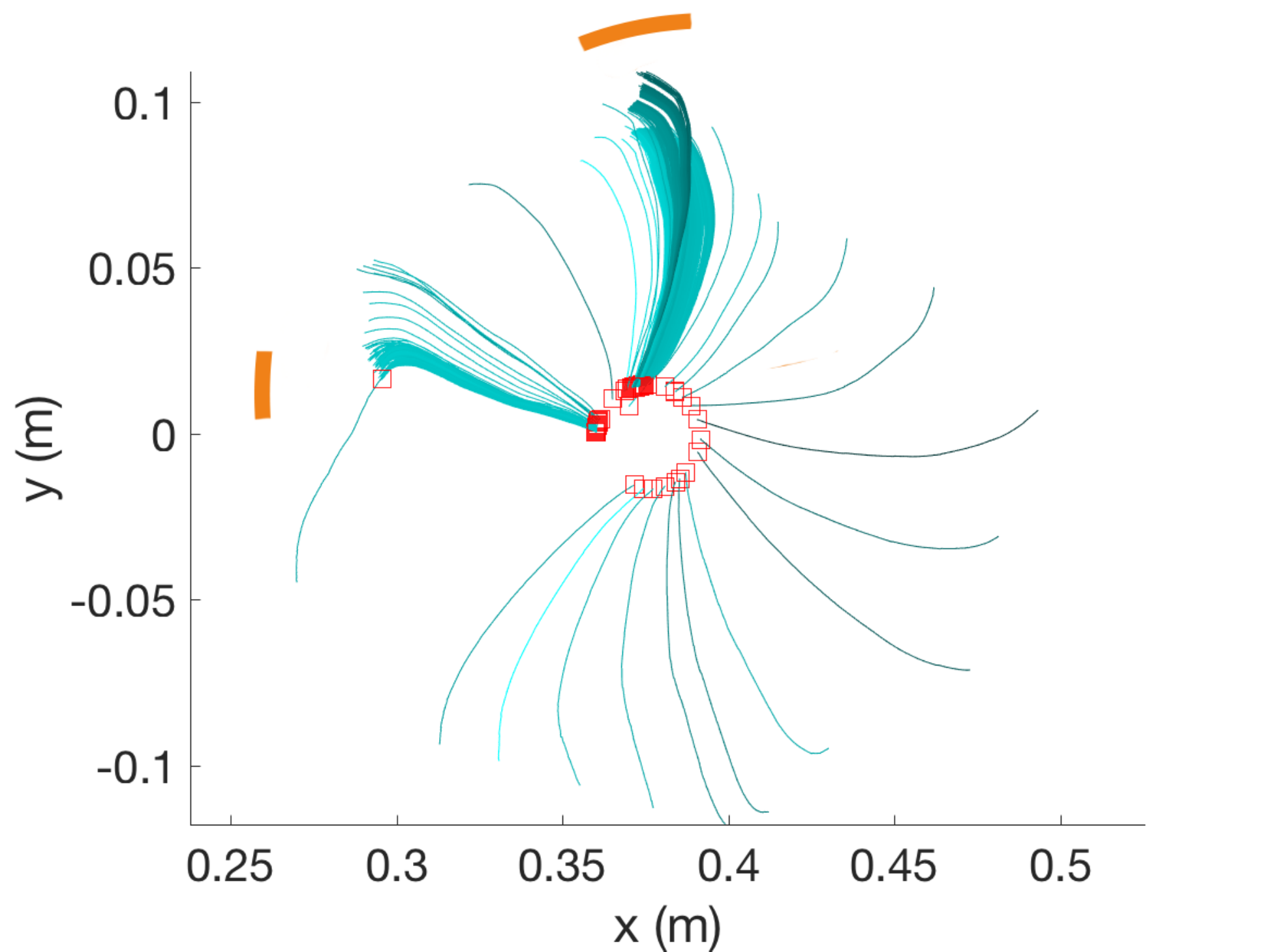}
    \end{minipage}
    
    \vspace{.25cm} 
    
    \begin{minipage}[t]{0.49\linewidth}
    \centering
    {delrin}
    \includegraphics[width=\linewidth]{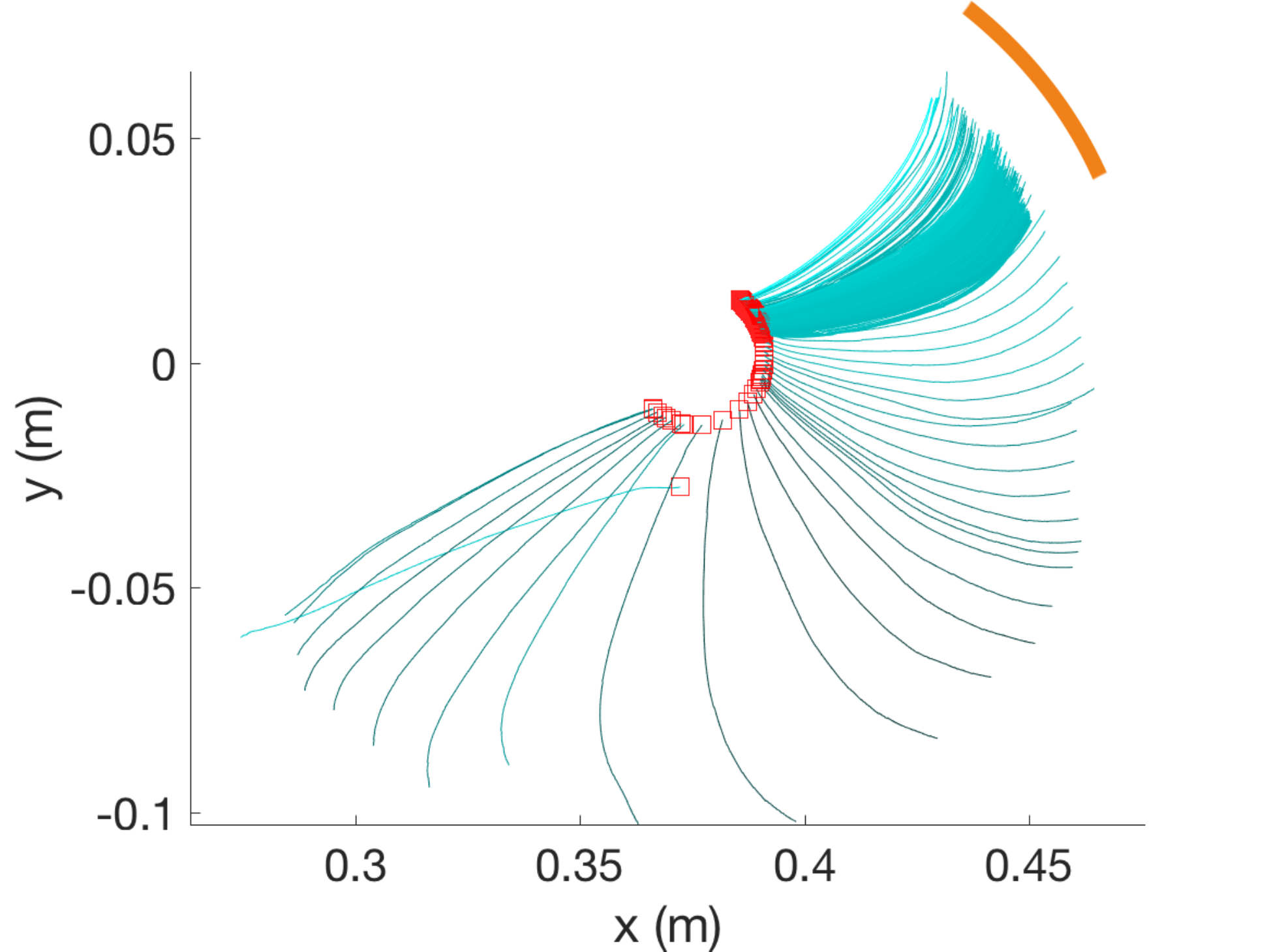}
    \end{minipage}
    \begin{minipage}[t]{0.49\linewidth}
    \centering
    {pu}
    \includegraphics[width=\linewidth]{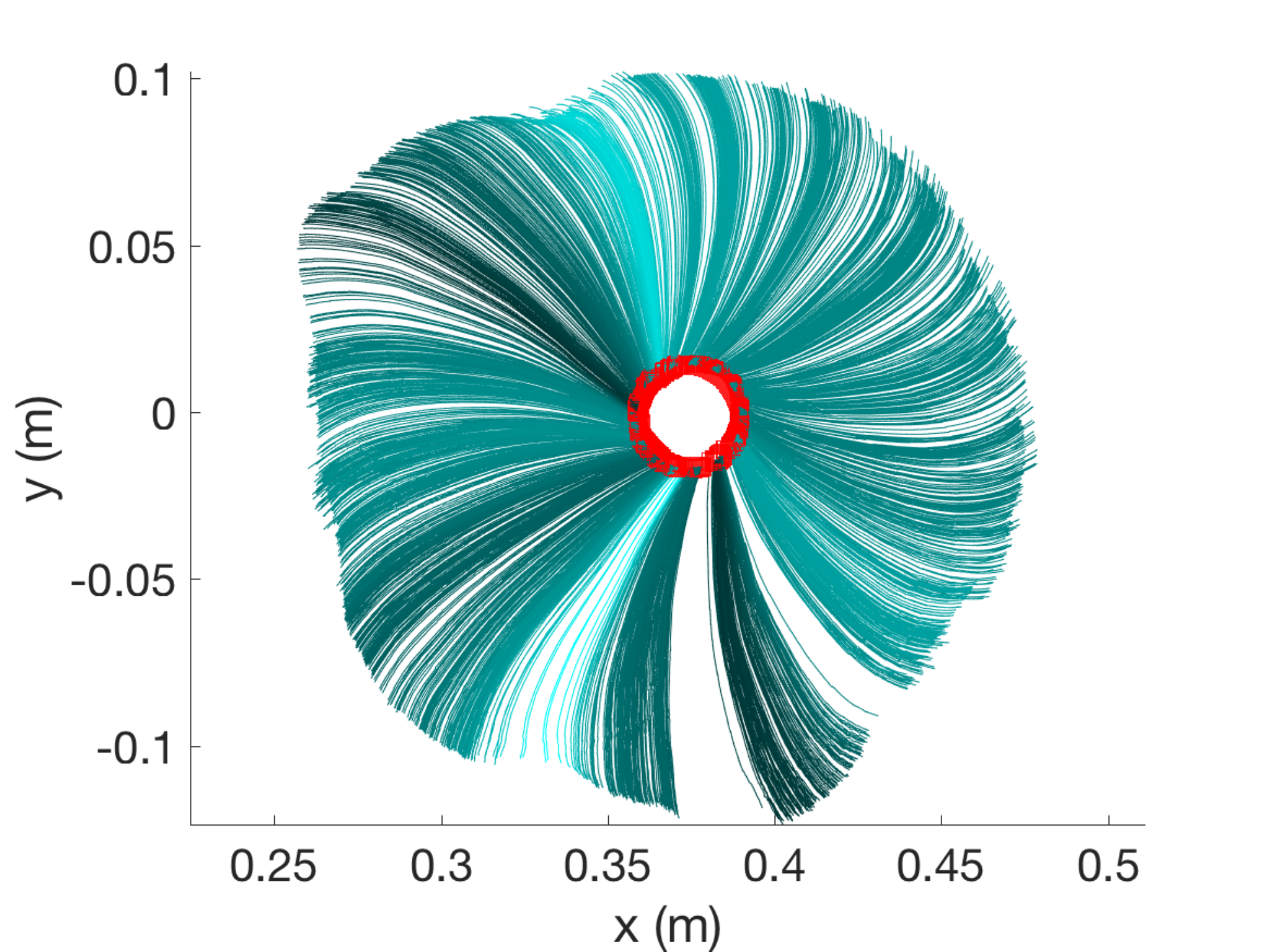}
    \end{minipage}
    \caption{Trajectories of the CoM of the pushed object in the global reference frame on four different materials: plywood, abs, delrin, and polyurethane. A darker shade represents a larger change in $\Delta\theta$. We label with orange marks the regions with higher density of trajectories.}
    \label{fig:Trajectories of CM in Robot Base}
\end{figure}

Figures~\ref{fig:Trajectories of CM in Object Base} and \ref{fig:Trajectories of CM in Robot Base} illustrate it.
\figref{fig:Trajectories of CM in Object Base} represents the output of experiments the way contact models are used in robotics (all position and direction dependence lumped in one single model). \figref{fig:Trajectories of CM in Robot Base} unwraps the trajectories in the global reference frame which is more informative.

Trajectories starting from different initial orientations bend in different directions, which suggests a marked anisotropy. %
More significantly, trajectories from different initial orientation in Fig.~\ref{fig:Trajectories of CM in Robot Base} seldom cross each other, which suggests a stable deterministic friction law between object and the surface. 

The common assumption that friction is uniform and isotropic leads to a dynamic system represented by the aggregated plot in the object frame Fig.~\ref{fig:Trajectories of CM in Object Base}. This shows artificial uncertainty due to "state compression". When we decompress the state in \figref{fig:Trajectories of CM in Robot Base}, we see that anisotropy is a likely cause. In \secref{sec:formation_uncertainty} we will analyze this relationship.

\begin{figure}[b]
    \begin{minipage}[t]{0.5\linewidth}
    \centering
    {plywood}
    \includegraphics[width=\linewidth]{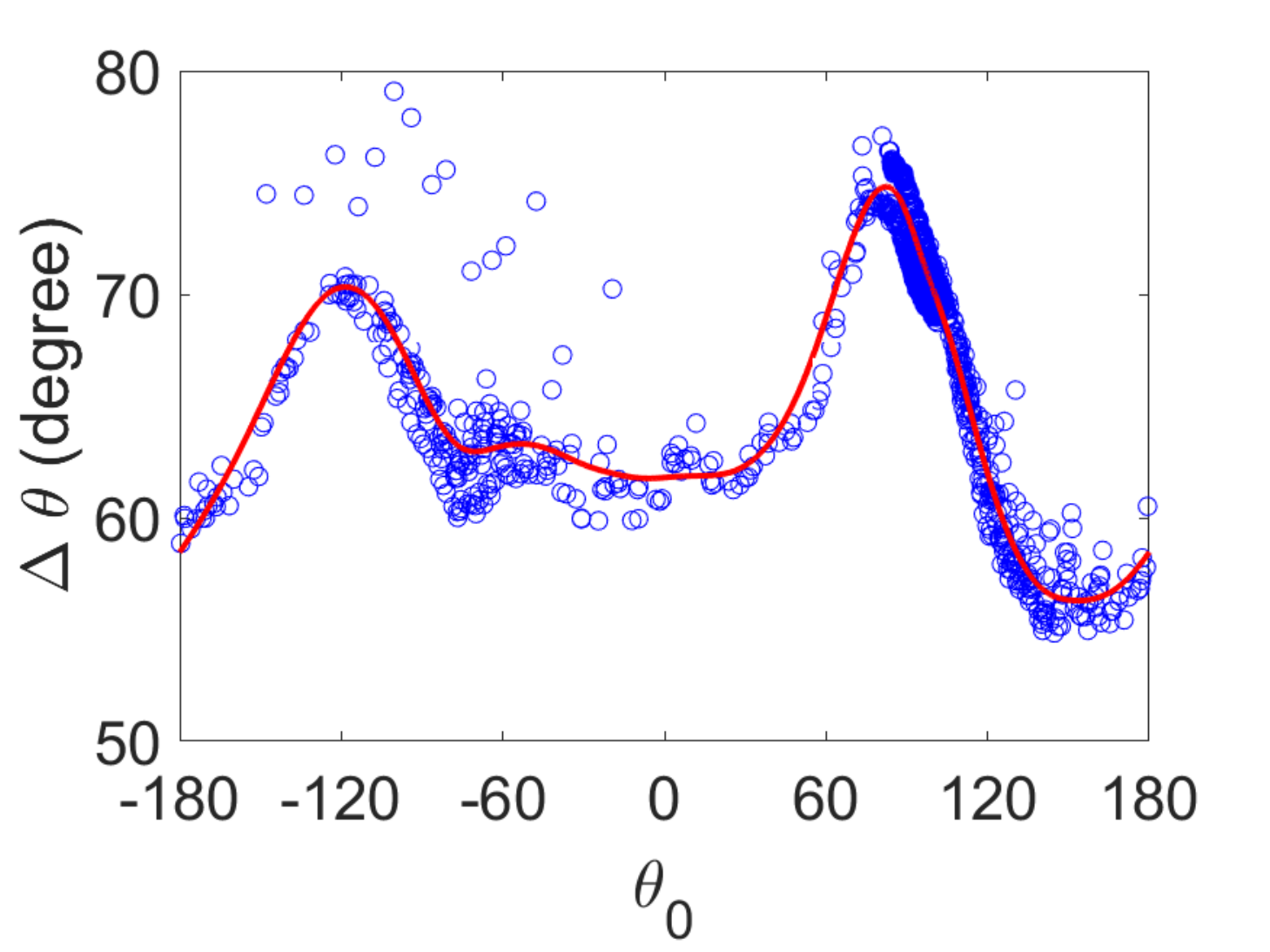}
    \end{minipage}%
    \begin{minipage}[t]{0.5\linewidth}
    \centering
    {abs}
    \includegraphics[width=\linewidth]{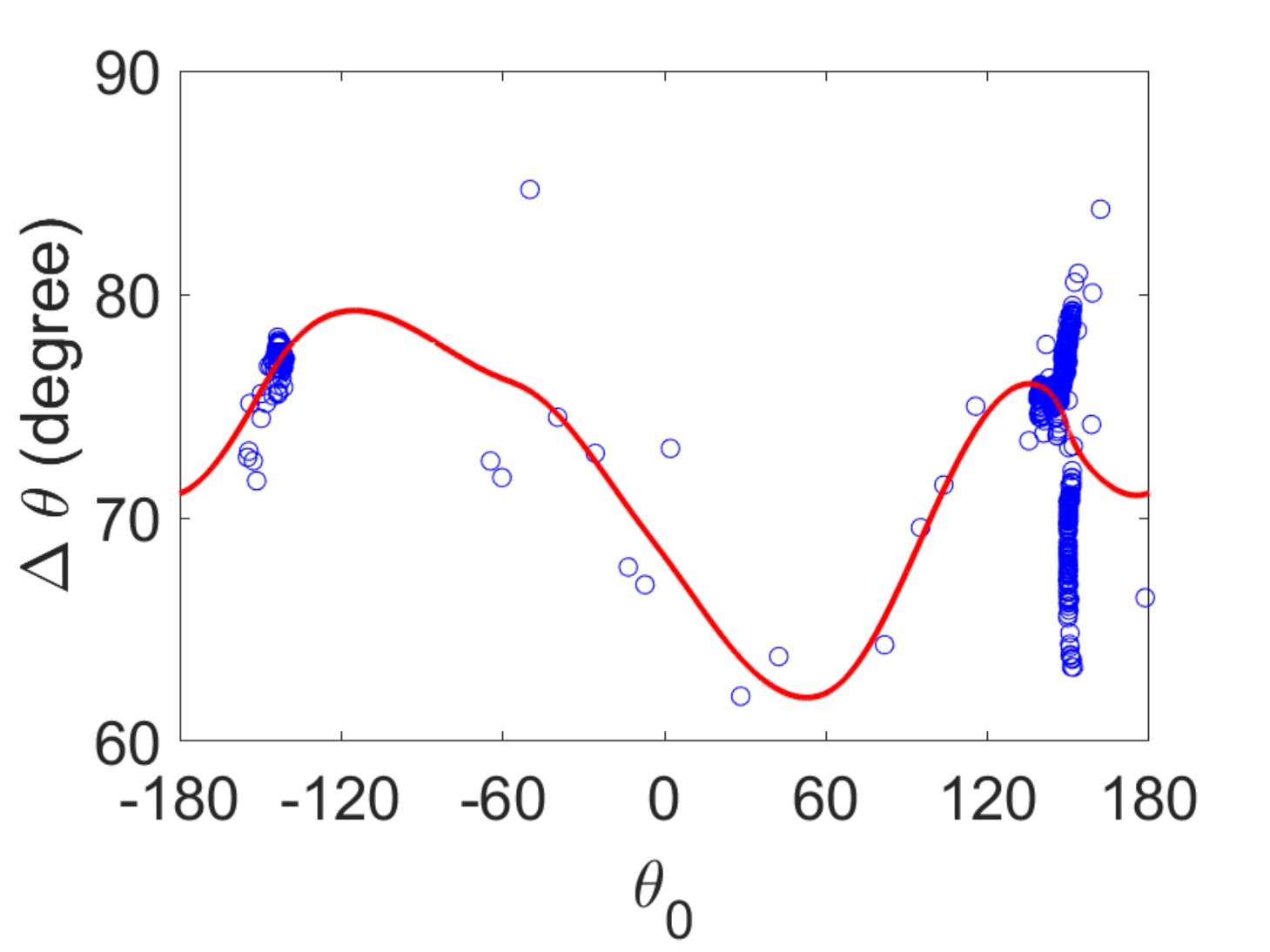}
    \end{minipage}
    
    \vspace{.1cm} 
    \begin{minipage}[t]{0.49\linewidth}
    \centering
    {delrin}
    \includegraphics[width=\linewidth]{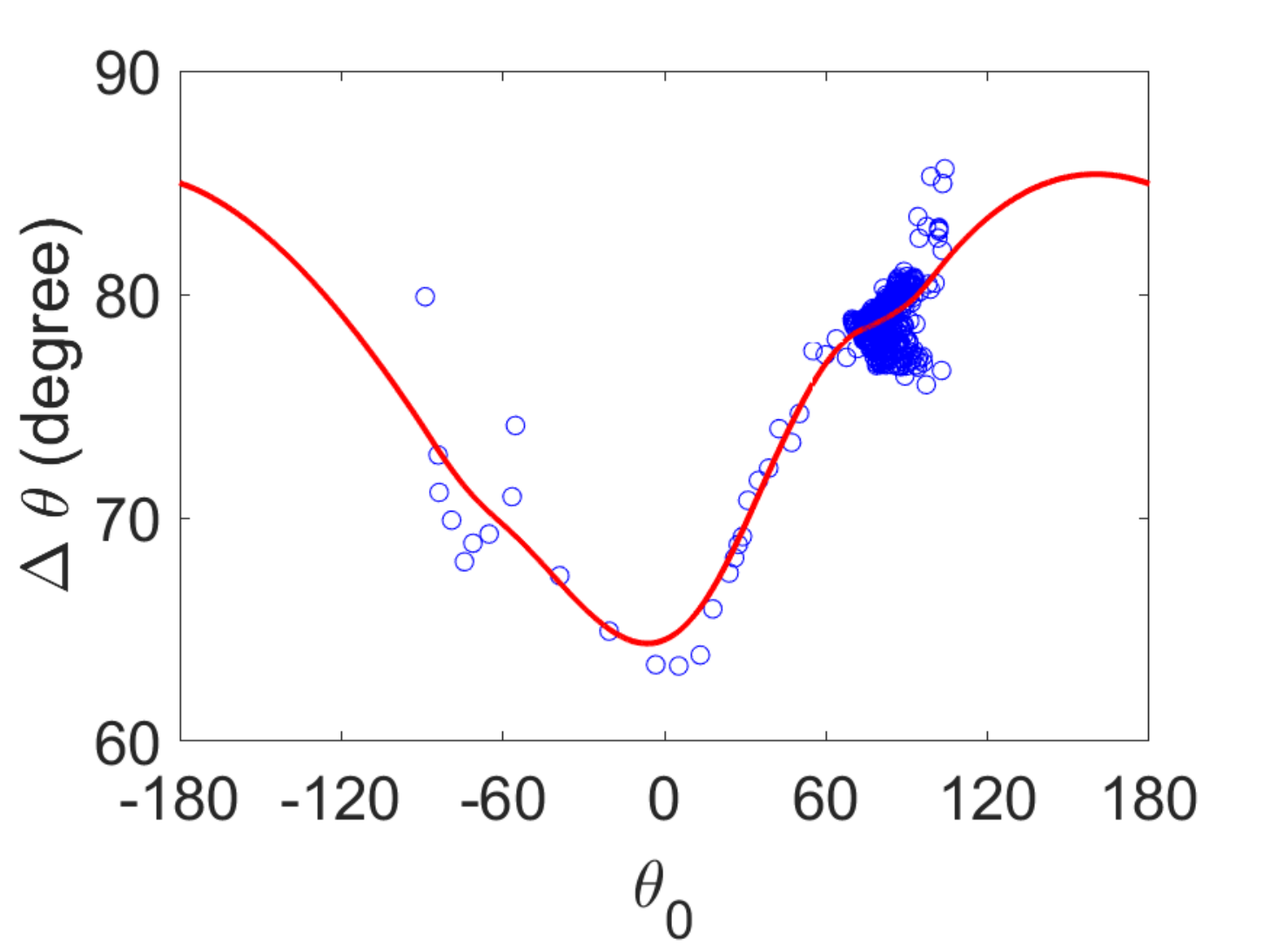}
    \end{minipage}
    \begin{minipage}[t]{0.49\linewidth}
    \centering
    {pu}
    \includegraphics[width=\linewidth]{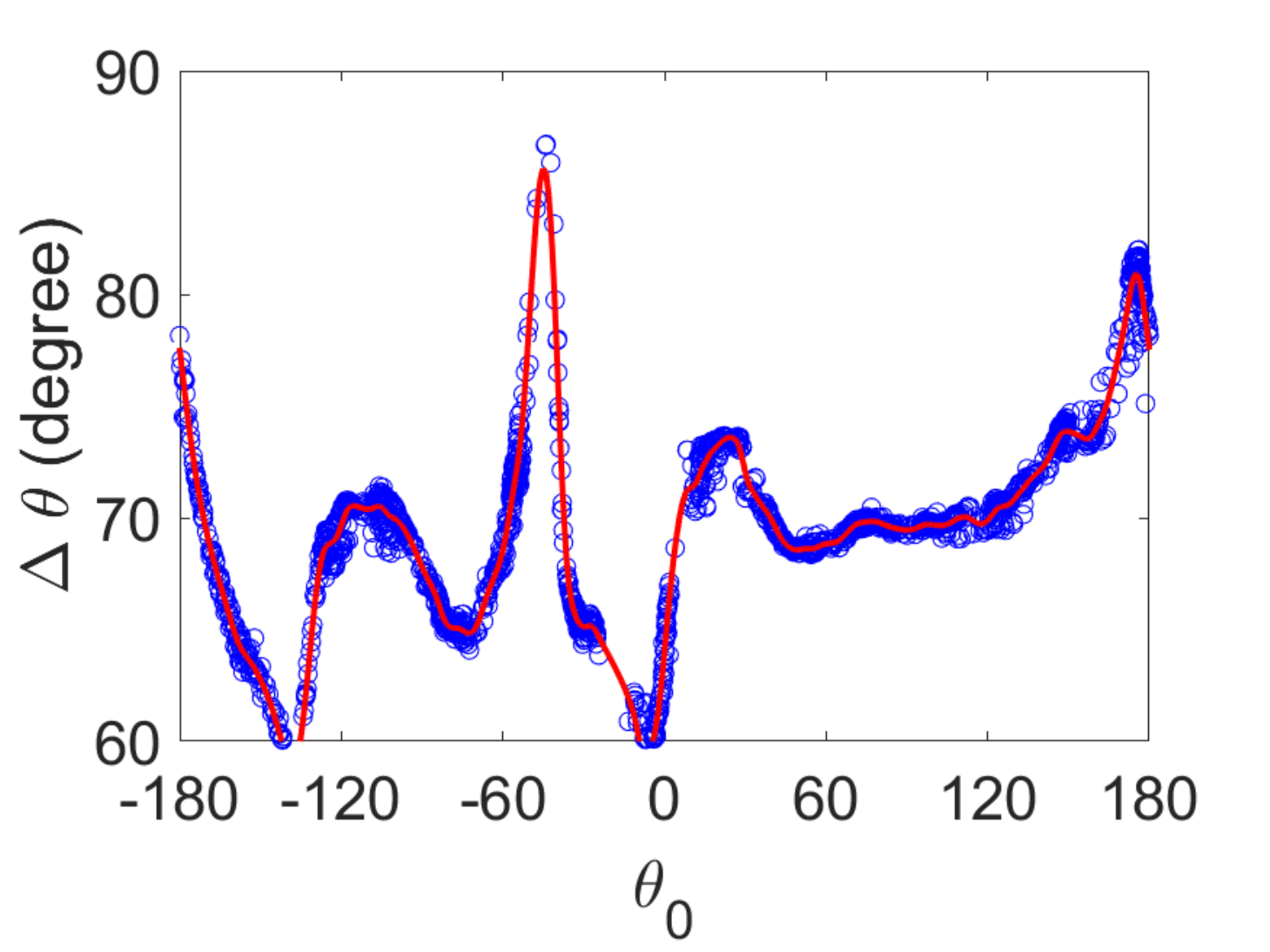}
    \end{minipage}
    \caption{Dependence of $\Delta\theta_0$ with the initial orientation $\theta_0$ of the object on four different materials. Red solid lines represents fitted curve while blue dots represent experimental data. 
    }
    \label{fig:Pushed angle versus initial orientation}
\end{figure}

\myparagraph{Stochasticity vs. determinicity}
\label{sec:determinicity}
%
The density plots in \figref{fig:Pushed angle versus initial orientation}, which show the change in orientation after a push as a function of the initial orientation, also suggest an almost deterministic relationship.
The plot shows that the dynamics of frictional pushing are quite stable given the initial orientation $\theta_0$, even after hundreds of pushes and re-positionings starting from the same initial orientation.
Although there is still some noise, the plots suggest a clear functional rather than statistical relationship between them. \change{The spread of data points on \textit{abs} and \textit{delrin} is likely due to wear and aging of material after hundreds of pushes. Friction shifts as experiment evolves. The material degradation for \textit{plywood} and \textit{pu} is not as severe since the pushes are less concentrated.}





\begin{figure}[b]
    \begin{minipage}[t]{0.5\linewidth}
    \centering
    {plywood}
    \includegraphics[width=\linewidth]{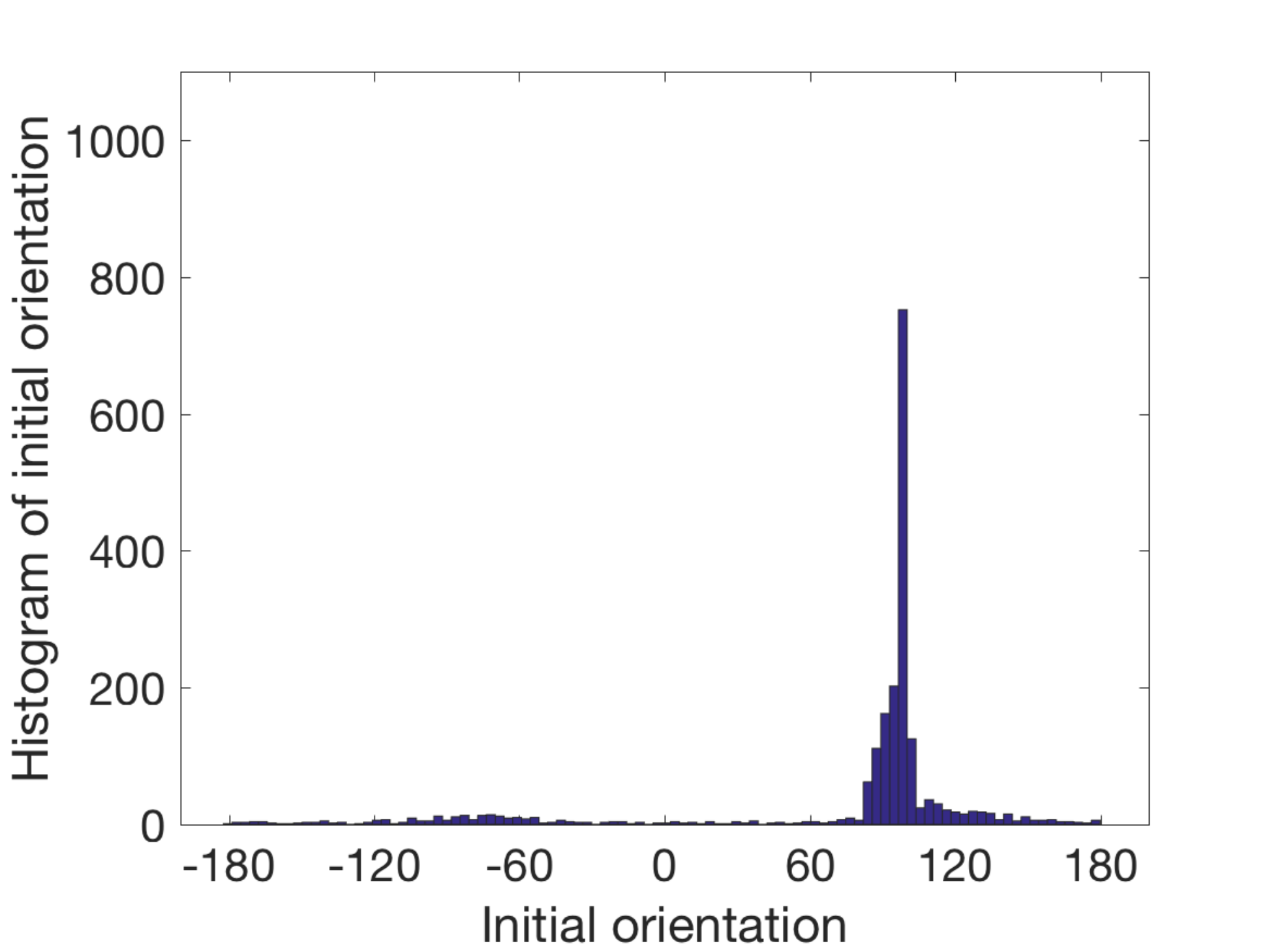}
    
    \end{minipage}%
    \begin{minipage}[t]{0.5\linewidth}
    \centering
    {abs}
    \includegraphics[width=\linewidth]{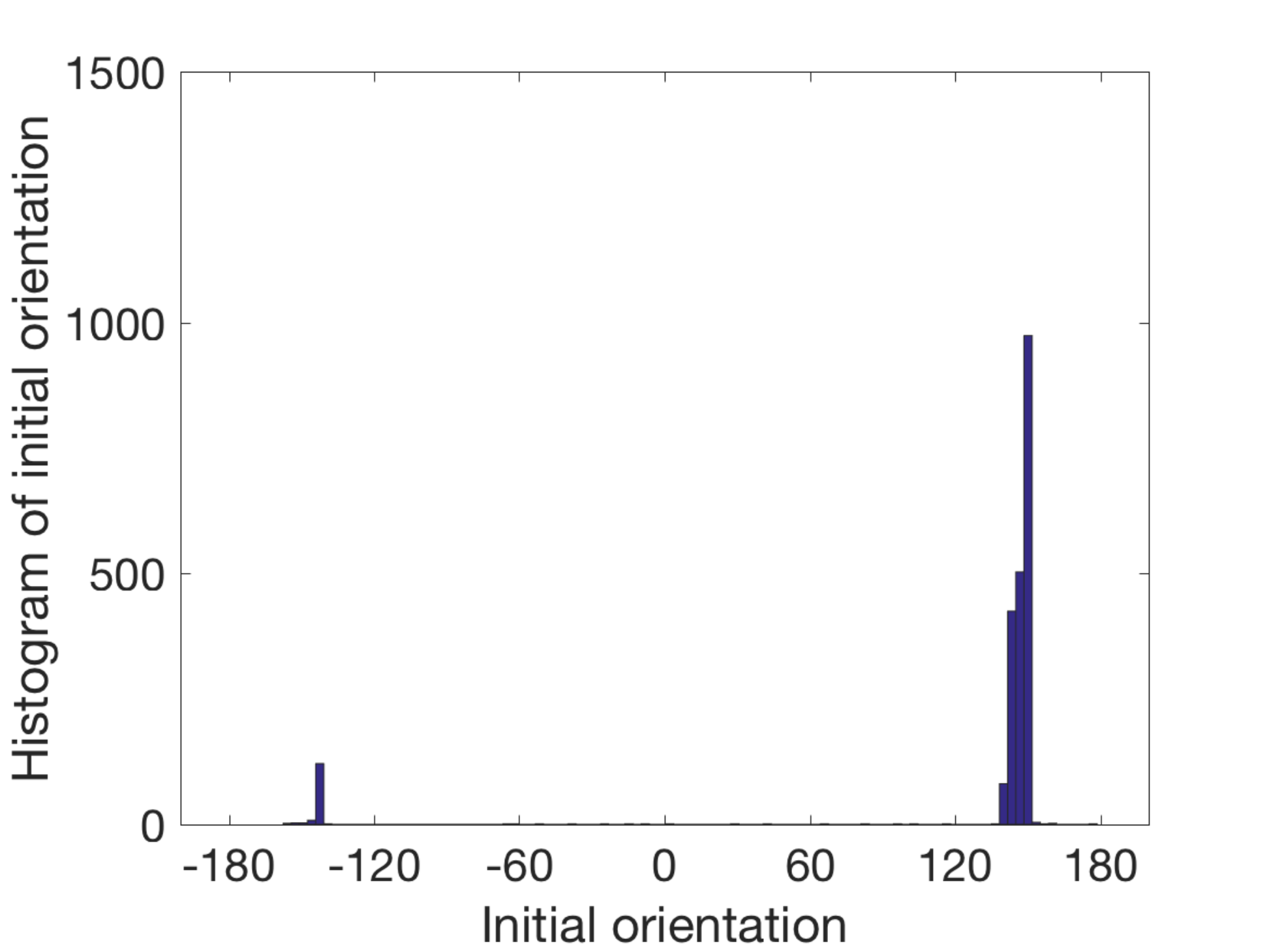}
    
    \end{minipage}
    
    \vspace{.15cm} 
    \begin{minipage}[t]{0.49\linewidth}
    \centering
    {delrin}
    \includegraphics[width=\linewidth]{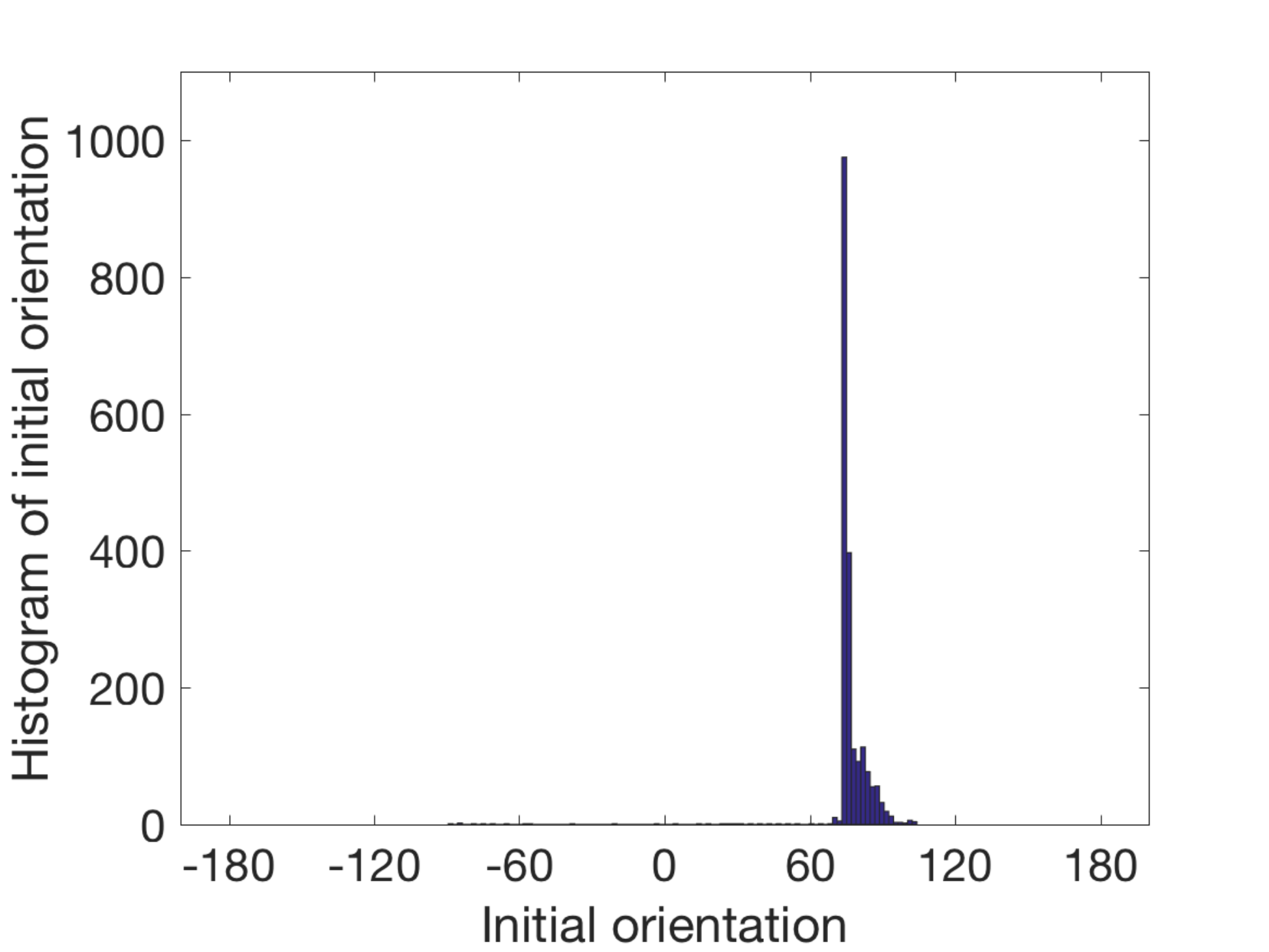}
    
    \end{minipage}
    \begin{minipage}[t]{0.49\linewidth}
    \centering
    {pu}
    \includegraphics[width=\linewidth]{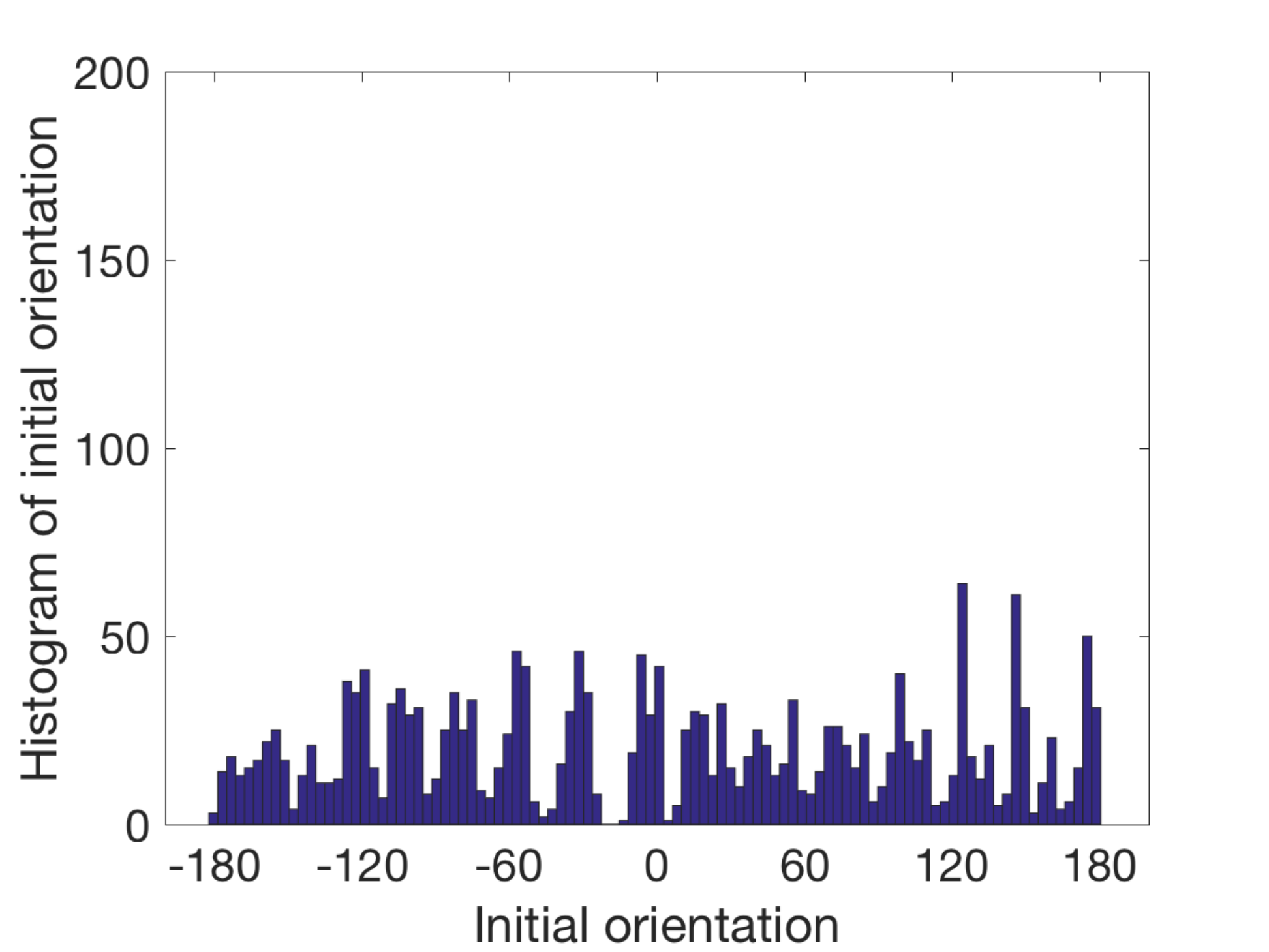}
    
    \end{minipage}
    \caption{Bias of initial orientation: Histogram of initial orientations in the dataset collected for four different materials.}
    \label{fig:IO histogram}
\end{figure}

\myparagraph{Bias in data-set}
Figure~\ref{fig:IO histogram} shows histograms for the initial orientation of the object for all four materials. These data, indicating strong bias in distribution, are very far from uniformly distributed and different from each other. 
On \textit{plywood}, we observe one high peak at around $100^{\circ}$ and a small one at its opposite direction. In \textit{abs} we only observe a small peak at around $-150^{\circ}$ and a high peak at $150^{\circ}$. On \textit{delrin} we only observe one peak at around $80^{\circ}$. We refer to these as stable directions. On the contrary, \textit{polyurethane} has a more uniform spectrum with multiple peaks at an almost periodic structure.

An initial clear observation is that the dataset is highly biased toward specific initial orientations.
In \secref{sec:formation_bias}, we discuss in depth how these distributions are formed, and why there are sharp peaks in some materials and not in others.

\subsection{Analysis: Formation of Structured Uncertainty -- Data Collection Dynamics}
\label{sec:formation_uncertainty}

In this part, we describe the structure of the variability/uncertainty of the pushing motions as a combination of three factors: a compressed state representation, bias of data-set, and anisotropic friction. 

For doing so, we start by attempting to recreate the histograms of the final poses of the pushed object by:
\begin{itemize}
    \item Drawing samples from the distribution of initial orientation of the object depicted in \figref{fig:IO histogram}.
    \item Use the fitted relationship in \figref{fig:Pushed angle versus initial orientation} to project the initial to final orientation of the object.
\end{itemize}

\begin{figure}[b]
    \begin{minipage}[t]{0.5\linewidth}
    \centering
    {plywood}
    \includegraphics[width=\linewidth]{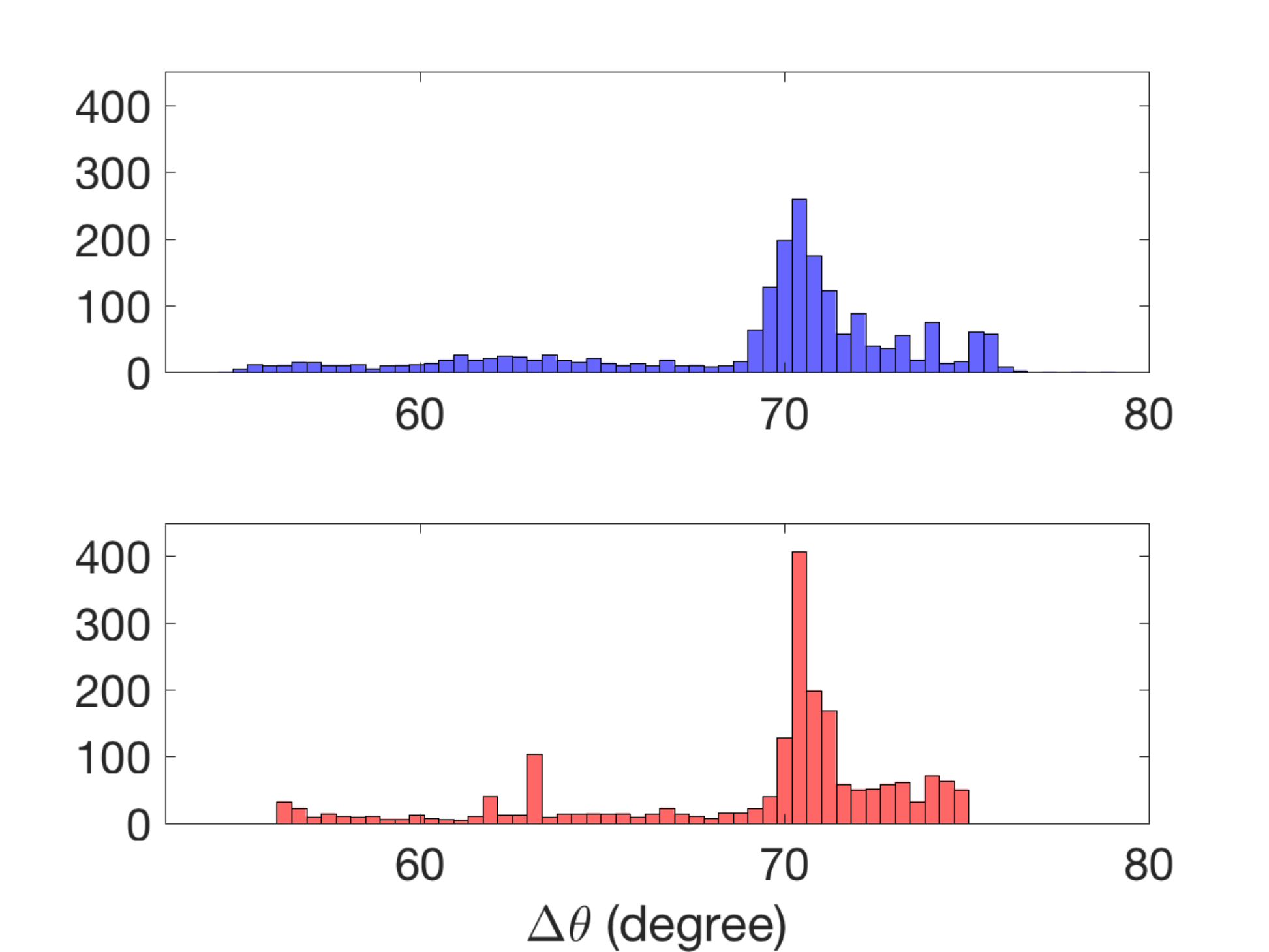}
    
    \end{minipage}%
    \begin{minipage}[t]{0.5\linewidth}
    \centering
    {abs}
    \includegraphics[width=\linewidth]{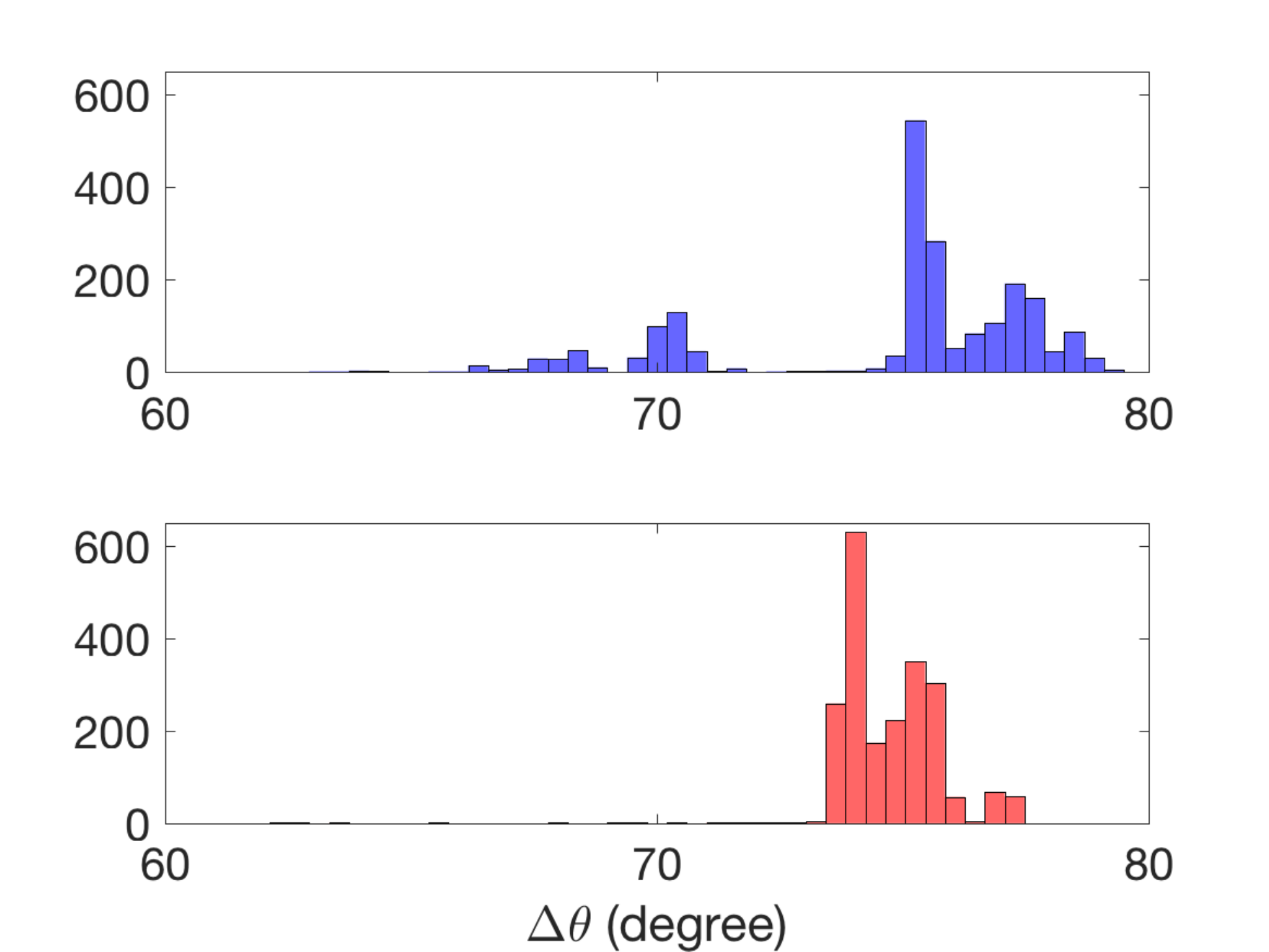}
    
    \end{minipage}
    
    \vspace{.15cm} 
    \begin{minipage}[t]{0.49\linewidth}
    \centering
    {delrin}
    \includegraphics[width=\linewidth]{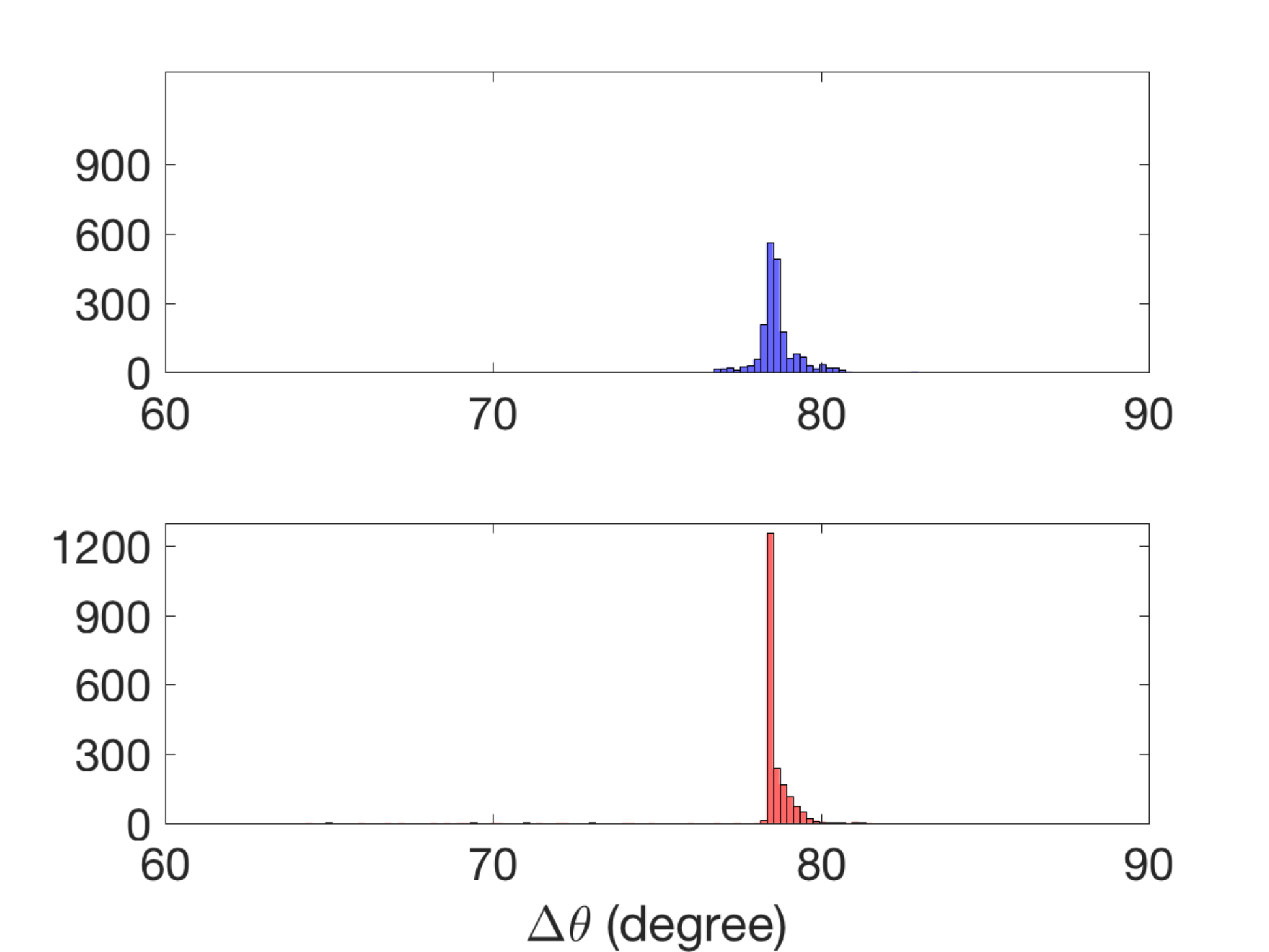}
    \end{minipage}
    \begin{minipage}[t]{0.49\linewidth}
    \centering
    {pu}
    \includegraphics[width=\linewidth]{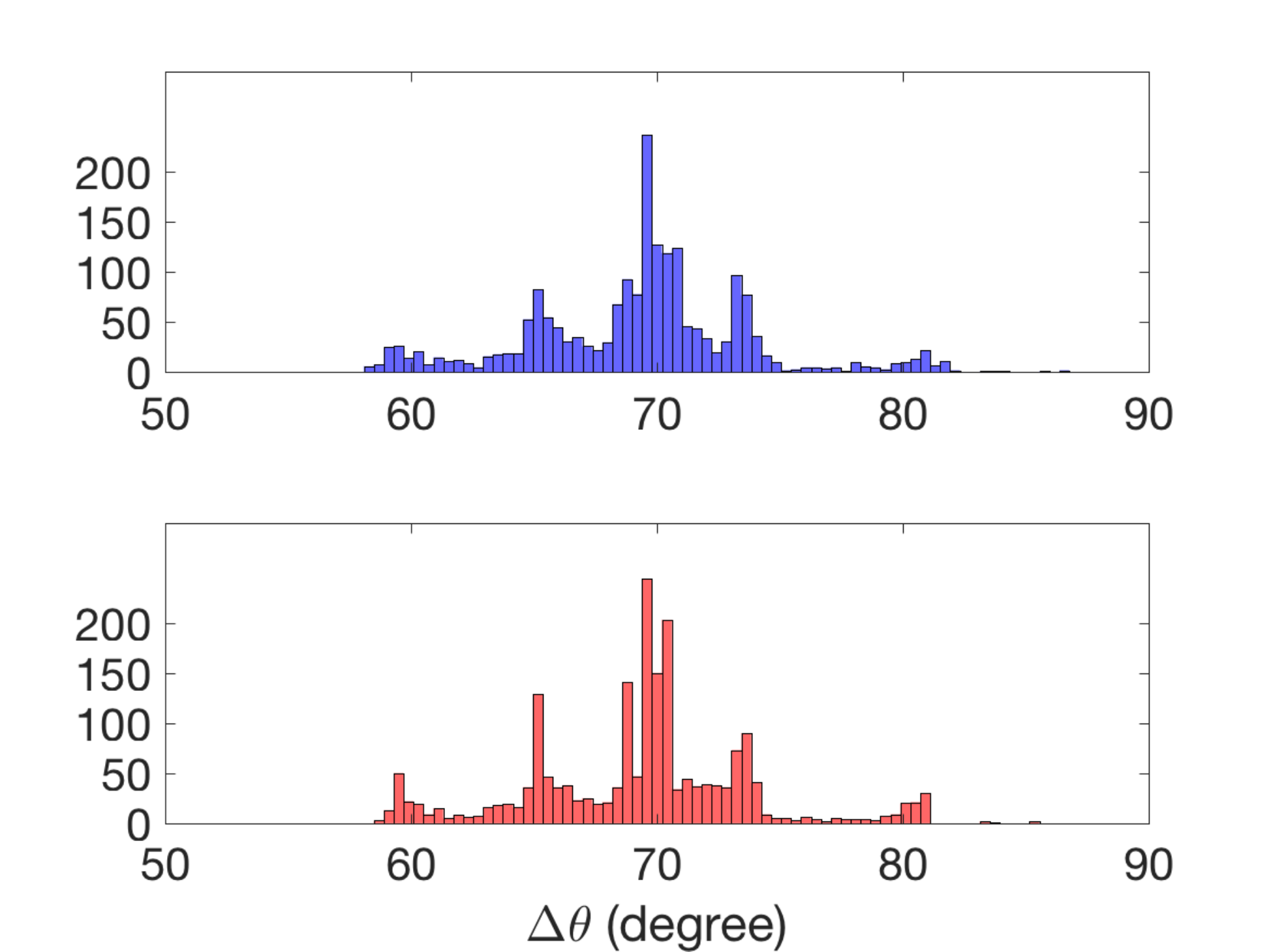}
    \end{minipage}
    \caption{\change{Histograms of $\Delta \theta$ on four different materials. Red bars represent prediction of 'fitted law', while blue bars represent experimental data.}}
    \label{fig:HistogramDelta_theta_COMPARE}
\end{figure}

The result is shown in \figref{fig:HistogramDelta_theta_COMPARE} which is strikingly similar to the real histograms for all four materials. The only exception is an extra lump in the histogram of \textit{abs} which is attributed to severe wear of the material after hundreds of pushes in the same area.

For numeric comparison, the root mean square error (RMSE) of prediction with an isotropic and homogeneous Coulomb friction and the fitted 'anisotropic law' are shown in \tabref{tab:isotropic} and Tables~\ref{tab:my_label1} respectively.
\tabref{tab:isotropic} shows the deviation of all the $\Delta\theta, \Delta x$ and $\Delta y$ in the data-set for a particular push. These numbers measure the variability of the data, and can be understood as the smallest RMSE an isotropic and homogeneous friction law can achieve at predicting it, which indicate the limitation of predictive performance by an isotropic and homogeneous friction law.

Tables~\ref{tab:my_label1} shows the RMSE and percentage errors for the prediction by the fitted 'anisotropic law'. Note that RMSE is the square root of the mean of the squared differences between predicted values and observed values. 
%
We see that the anisotropic law with the biased initial conditions yields a much better fit to the data in all prediction aspects. Note also that the prediction error is considerably smaller for materials \textit{plywood} and \textit{pu}.

\begin{table}
    \centering
    \begin{tabular}{c|c c |c c  |c c}
    \hline
    $\sigma$    & $\Delta\theta $ & \% & $\Delta x$ & \% & $\Delta y$& \%  \\
        & (degree)  &  & (mm) &     & (mm) &   \\
    \hline
\rowcolor[gray]{0.8}     plywood        & 4.74   &  6.86  &   2.84  &   7.75 &    8.57  &   9.09\\
\rowcolor[gray]{0.8}     abs            & 3.21   &  4.30  &   1.59  &   3.97 &    5.31  &  7.86\\
\rowcolor[gray]{0.8}     delrin         & 1.28   &  1.63  &   0.97  &   2.49 &    3.21   &  6.33\\
\rowcolor[gray]{0.8}     pu             & 4.53   &  6.53  &   2.93  &   7.30 &   11.39   & 13.40\\
\hline
    \end{tabular}
    \caption{Standard deviation of $(\Delta\theta, \Delta x,\Delta y)$ normalized by the mean of $(\Delta\theta, \Delta x,\Delta y)$.}
    \label{tab:isotropic}
\end{table}

\begin{table}
    \centering
    \begin{tabular}{c|c  c |c c| c c}
    \hline
    RMSE    & $\Delta\theta $ & \% & $\Delta x$ & \% & $\Delta y$& \%  \\
        & (degree)  &  & (mm) &     & (mm) &   \\
    \hline
   \rowcolor[gray]{0.8} plywood                         & \change{2.07}   &   \change{2.99}   & \change{0.91}    & \change{2.46}     & \change{3.39}   &  \change{3.66}  \\

   \rowcolor[gray]{0.8} abs                             & \change{2.99}   &  \change{4.00}    & \change{1.56}    &  \change{3.88}    & \change{3.21}     &  \change{4.75} \\
 
   \rowcolor[gray]{0.8} delrin                          & \change{1.28}   & \change{1.63}     & \change{0.81}    &  \change{2.07}    & \change{1.61}      &   \change{3.08} \\

   \rowcolor[gray]{0.8} pu                              & \change{1.14}   &  \change{1.64}    & \change{0.60}    &  \change{1.49}    & \change{1.56}       &   \change{1.86} \\
 
\hline
    \end{tabular}
    \caption{RMSE of  $(\Delta\theta, \Delta x,\Delta y)$ corresponded to fitted deterministic law, normalized by mean of $(\Delta\theta, \Delta x,\Delta y)$.}
    \label{tab:my_label1}
\end{table}


\subsection{Analysis: Formation of Data-set Bias}
\label{sec:formation_bias}

Recall that the histograms in \figref{fig:IO histogram} show a marked bias in the initial orientation of the object, implying that the experiments happen much more frequently along certain directions. We have seen that bias plays a key role in forming the artificial multi-modal structure in the histogram of final pushed motions $(\Delta x, \Delta y, \Delta\theta)$. In this section we study how that bias is formed.


To characterize the bias, we need to better understand the data collection dynamics by analyzing the time history of initial orientations for data collection. As illustrated in \figref{fig:Illustration_of_data_collection_dynamics}, the position and initial orientation at push $(k+1)$ differs from that at push $k$. These are related through the cyclic dynamics of pushing and dragging back.


The curves in \figref{fig:Pushed angle versus initial orientation} show the change in orientation after the pushing phase. Now we are interested in the change in orientation after the combination of pushing and dragging phases.
Let $f(\theta_0^k) = \theta_0^{k+1}$ be the function that maps an initial orientation to the next initial orientation. 

Figure~\ref{fig:Accumulation of pushed angle} shows the evolution of $\theta_0$ as the experiment progresses, without removing the natural $2\cdot\pi$ winding of the angle. Each material shows a slightly different behavior:

\begin{figure}[t]
    \begin{minipage}[t]{0.5\linewidth}
    \centering
    {plywood}
    \includegraphics[width=\linewidth]{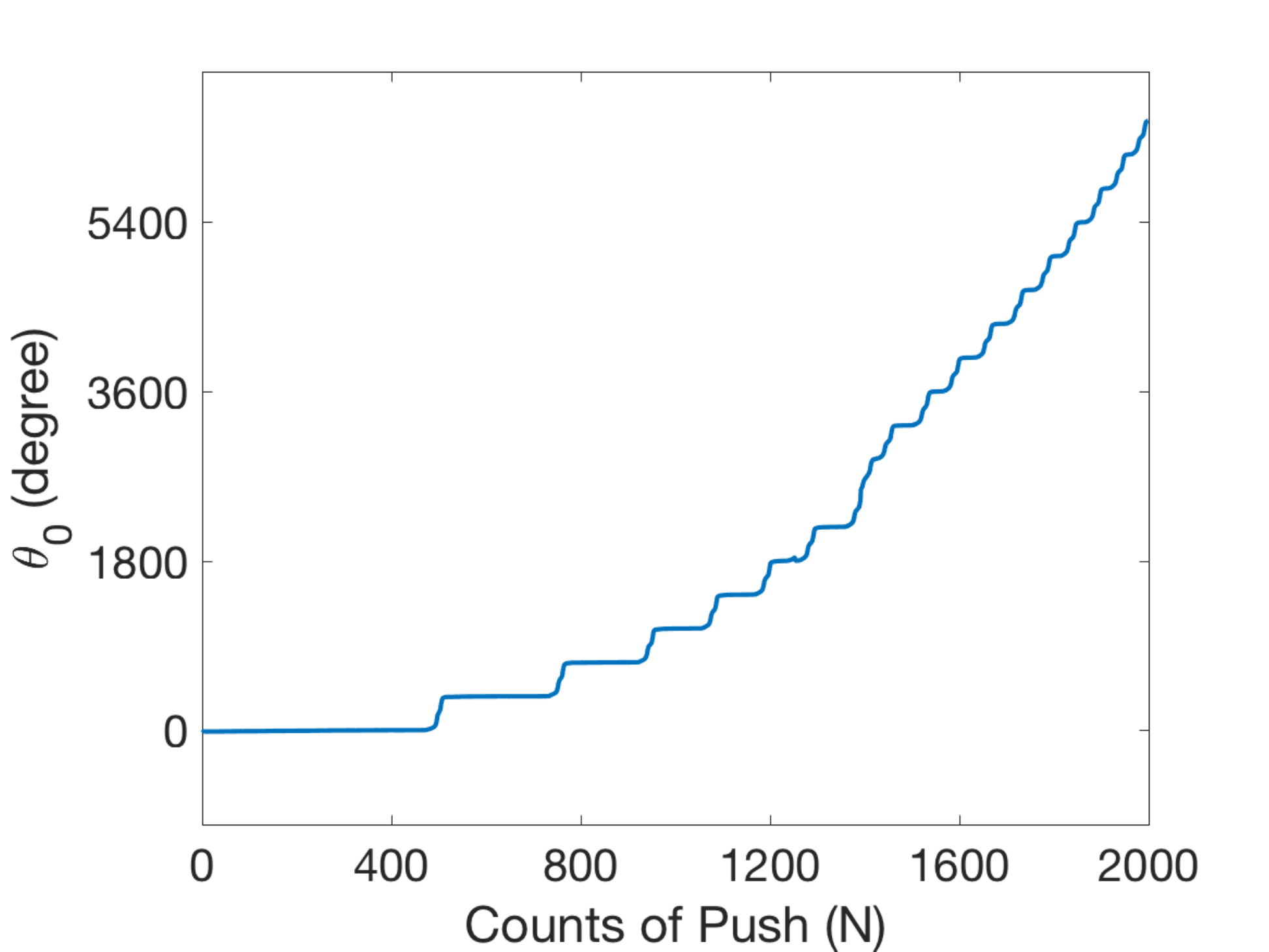}
    \end{minipage}%
    \begin{minipage}[t]{0.5\linewidth}
    \centering
    {abs}
    \includegraphics[width=\linewidth]{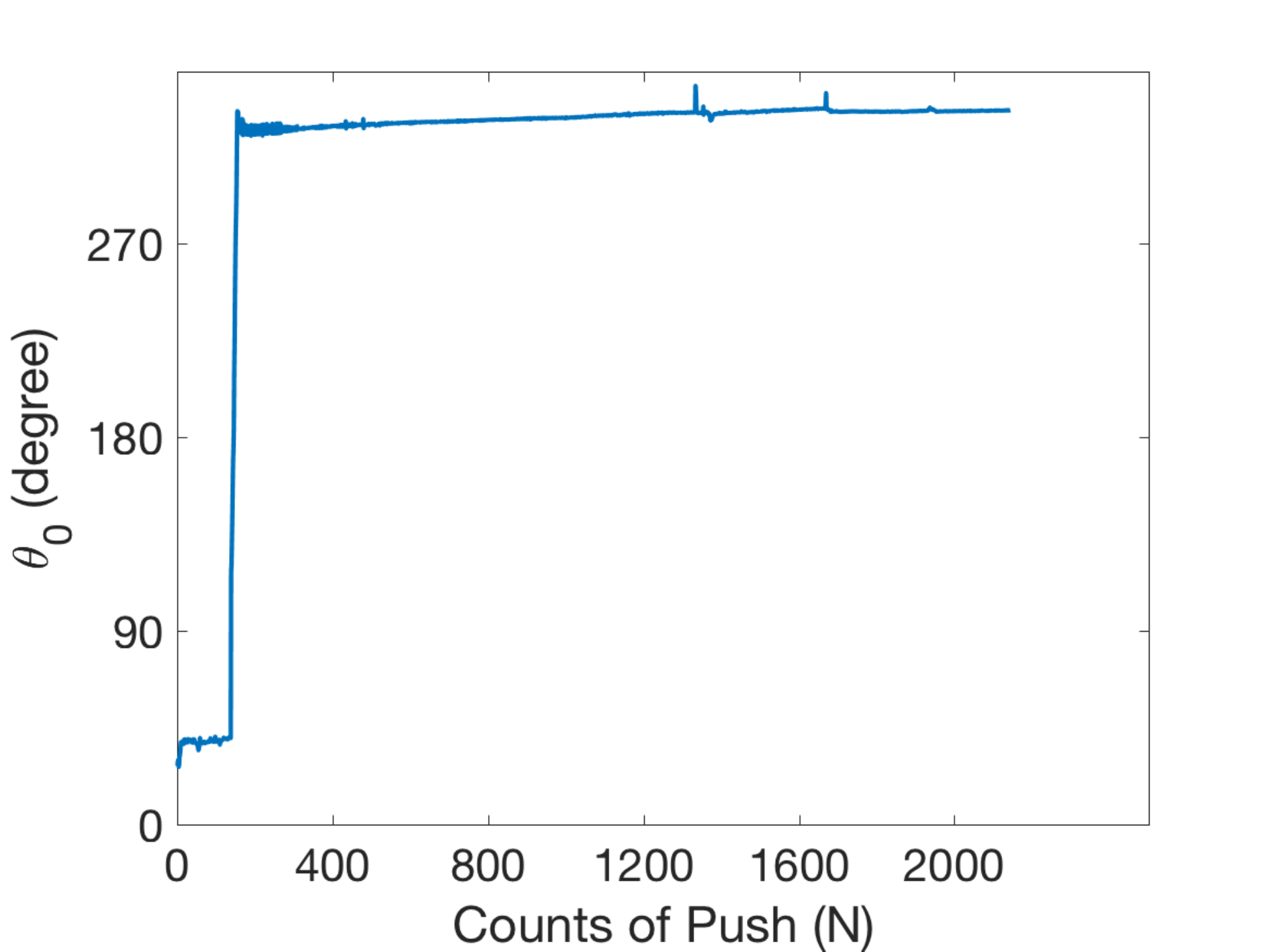}
    \end{minipage}
    \begin{minipage}[t]{0.48\linewidth}
    \centering
    {delrin}
    \includegraphics[width=\linewidth]{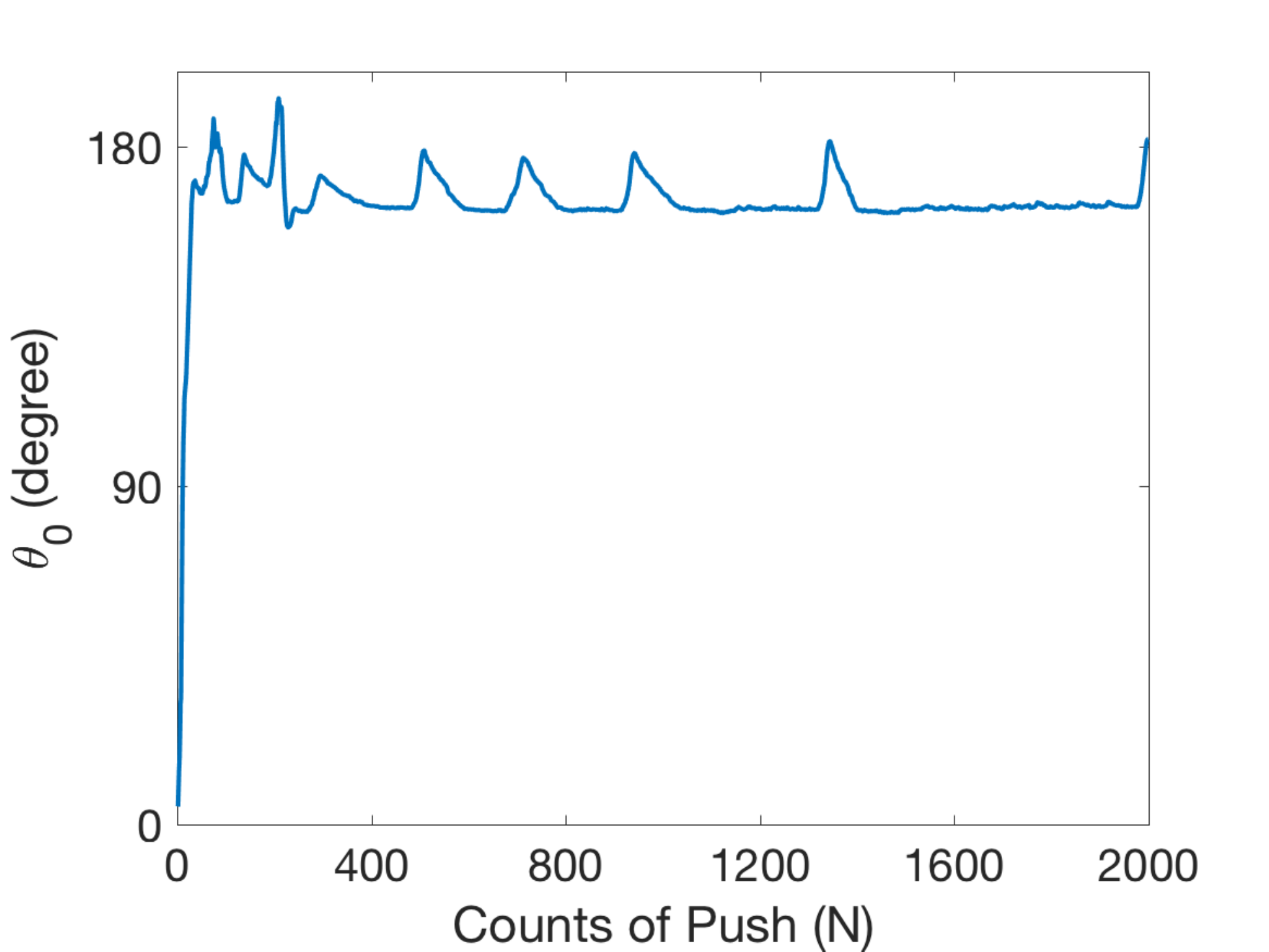}
    \end{minipage}
    \begin{minipage}[t]{0.5\linewidth}
    \centering
    {pu}
    \includegraphics[width=\linewidth]{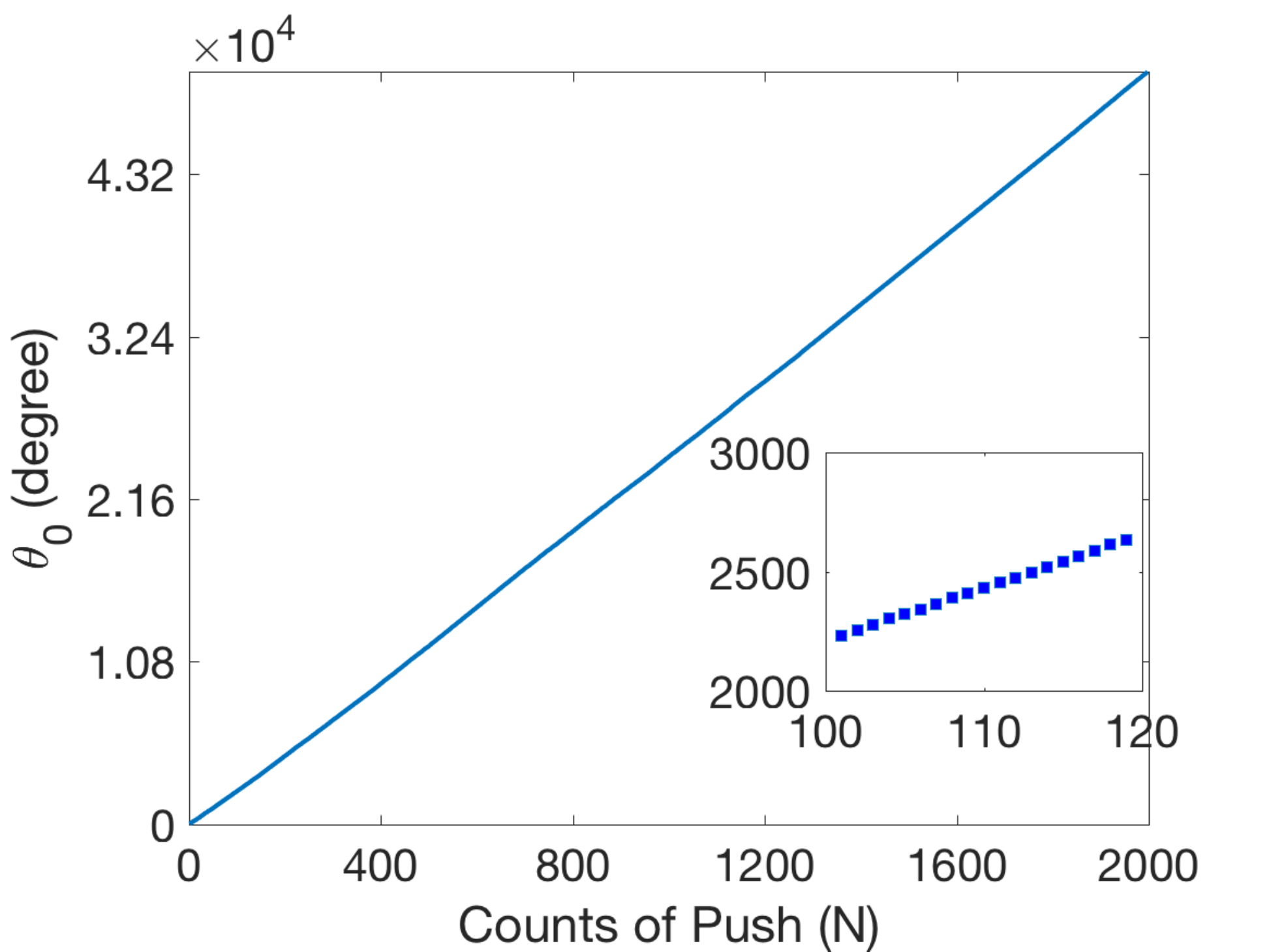}
    \end{minipage}
    \caption{Evolution of the initial orientation of pushed object $\theta_0$ during 2000 consecutive pushes on four different materials.}
    \label{fig:Accumulation of pushed angle}
\end{figure}

\begin{itemize}
\item \textbf{plywood}: The curve shows a stair-like growth with a step height of exactly $360^{\circ}$. The flatness of the step corresponds to the stable direction where experiments repeats. A step height of $360^{\circ}$ implies that there is only one stable direction for the initial orientation. Signs of wear on \textit{plywood} are also apparent, since the "stability" of that direction seems to decrease as the experiment progresses. 

\item \textbf{abs}: The initial orientation gets "stuck" in one stable direction where experiments repeat for about 150 times. Then it escapes and gets trapped in second stable direction from where it never escapes again. This indicates that at least two stable directions exists for the data collection dynamics on \textit{abs}.

\item \textbf{delrin}: The evolution of the initial orientation is quite similar to that of \textit{abs}, with a possibly less stable orientation.

\item \textbf{pu}: Unlike all other materials, we do not observe any stable orientation in polyurethane. The cumulative initial orientation is always increasing 
\change{A zoomed-in view of \textit{pu} in \figref{fig:Accumulation of pushed angle} shows how the initial orientation changes very regularly versus the pushing count.}
\end{itemize}

It is no surprise that the stable directions in \figref{fig:Accumulation of pushed angle} correspond directly to the histogram peaks in \figref{fig:IO histogram}. It is key then to understand how the stable directions are formed. To answer this question, we propose an anisotropic friction model in \secref{sec:anisotropic} and use it to simulate the data collection process in \secref{sec:simulation}.

\section{Anisotropic friction law}
\label{sec:anisotropic}
The analysis in \secref{sec:determinicity} indicates a deterministic relationship between the initial and final orientation of the object, which suggests a deterministic friction law. Anisotropic friction is a good candidate because of its dependence with the sliding direction.


\begin{figure}[t]
    \centering
    \includegraphics[width=0.85\linewidth]{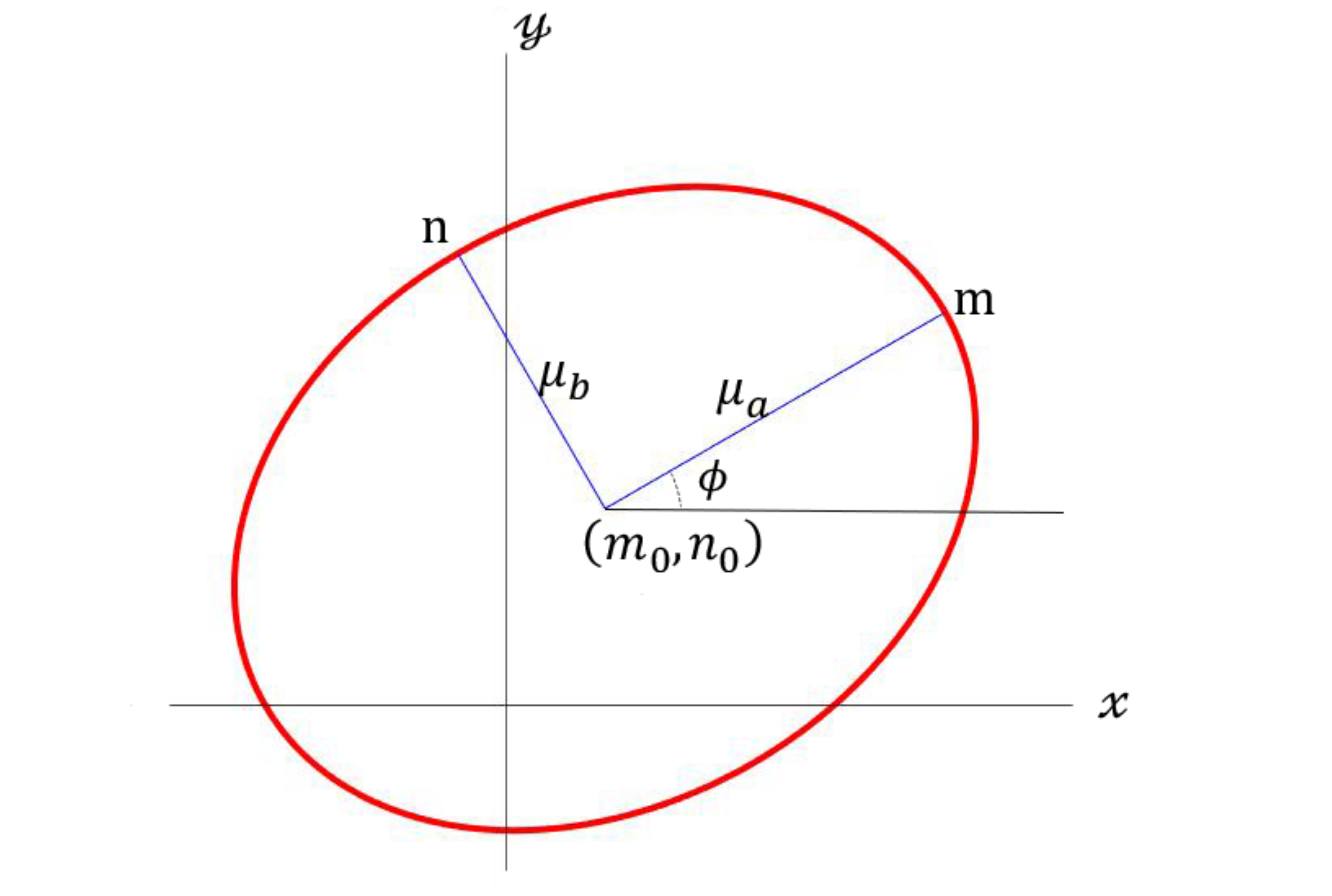}\\
    \vspace{0.05in}
    \includegraphics[width=0.75\linewidth]{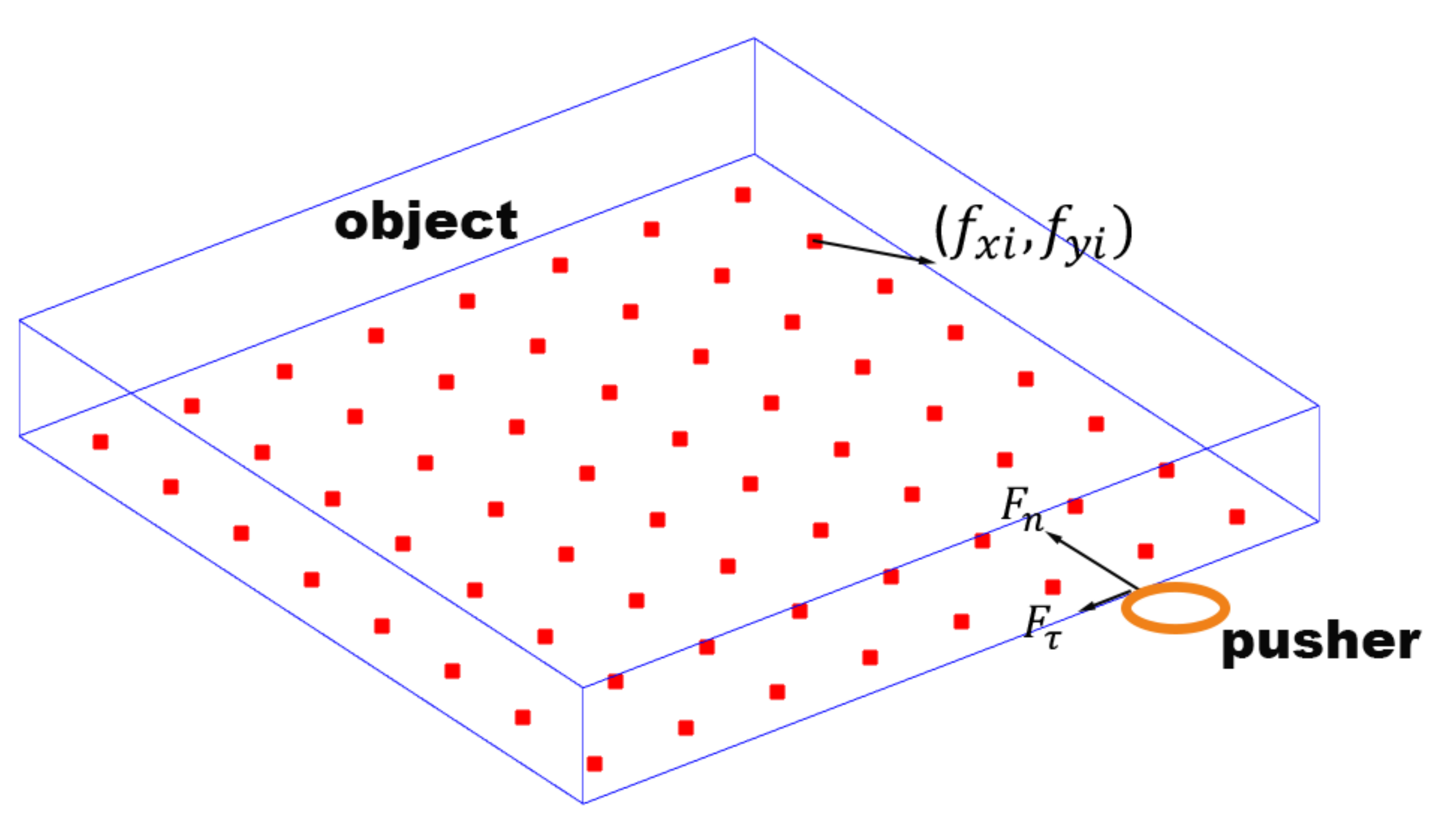}
    \label{fig:AnisotropicFrictionModelling}
    \caption{ \change{(\textbf{Top}) Elliptical friction limit circle for a single point contact between object and plywood. (\textbf{Bottom}) The contact patch between object and plywood is modeled as sets of rigidly connected point contacts. Friction forces at each point follows to the same friction law described by the ellipse limit circle on the left.}}
\end{figure}

Coulomb' friction law on a point contact can be represented as a circle of radius $\mu$, the \textit{limit circle}~\citep{Goyal1991PlanarFunction}. A natural generalization to include anisotropy is to define an equivalent elliptic limit circle (\figref{fig:AnisotropicFrictionModelling} (Left)).
In the general case, the center of ellipse is offset from the origin and the two principle axes are not parallel to $\bm{x}$ and $\bm{y}$ axis, and are of different magnitude.
We note the center of ellipse with $(m_0,n_0)$ and the two principle axes with $\bm{m}$ and $\bm{n}$, rotated by $\phi$ from $\bm{x}$ and $\bm{y}$. 

\change{We choose this model for its compactness as well as with a physical basis. Take the example of wood. Its texture represents the orientation of organized wood fibers. Although wood is usually viewed as bulk material, 
the micro-structure of wood fibers can yield anisotropic frictional behavior. If the wood fibers are parallel to each other, it is natural to expect two different coefficients for the two orthogonal directions following the fibers and orthogonal to them. Furthermore, while we usually think of texture as a 2D distribution, under the microscope, its micro-structures are 3D ridges. If these are not symmetric, sliding along opposite directions can produce different friction forces. This will produce a limit ellipse with an offset center.}


If we denote $(\mu_x,\mu_y)$ to be the coefficient of friction in x and y direction, then $\mu_m = \mu_x \cos\phi + \mu_y \sin\phi $ and $\mu_n = -\mu_x \sin\phi+ \mu_y \cos\phi$. Thus, the ellipse limit circle of anisotropic friction can be expressed as:
\begin{equation}
\frac{ (\mu_m  - m_0)^2 }{\mu_a^2} + \frac{(\mu_n - n_0)^2}{\mu_b^2} = 1;
\label{eq:1}
\end{equation}

If we denote the non-zero sliding velocity of a point contact with anisotropic friction to be $\bm{v}=v_x \bm{x} + v_y \bm{y} = v_m \bm{m} + v_n \bm{n}$, the maximum-power inequality, a.k.a. maximum dissipation principle, leads to a friction coefficient vector $\bm{\mu} = \mu_m \bm{m} + \mu_n \bm{n}$, where
\begin{equation}
\left\{
\begin{aligned}
\mu_m =~   & m_0 + \mu_a \frac{\mu_a v_m}{\sqrt{\mu_a^2 v_m^2 + \mu_b^2 v_n^2 }}\\
\mu_n =~  & n_0 + \mu_b \frac{\mu_b v_n}{\sqrt{\mu_a^2 v_m^2 + \mu_b^2 v_n^2 }}
\end{aligned}
\right.
\label{eq:2}
\end{equation}

The friction force is $\bm{f_\tau} = \bm{\mu} N$, where $N$ is the normal load. Since \change{\eref{eq:2}} is nonlinear on $v_m$ and $v_n$, one essential aspect of this anisotropic friction law is that the friction force $\bm{f_\tau}$ is not necessarily co-linear with the sliding velocity $\bm{v}$. The non-linearity and directional dependence of frictional force from anisotropic friction law complicates the pushed motion of the object and in consequence forms directional preferences of experiment distribution.

\section{Simulation}
\label{sec:simulation}

One of the key experimental observations in \secref{sec:facts} is that there are directional preferences for experiments and biases in distribution of initial condition.
As indicated in \secref{sec:determinicity}, anisotropic friction is a likely source of the viability. In order to reproduce the experimental phenomenon and validate the claim, we conduct numerical simulations of the data collection dynamics. For simplicity, we only carry out the simulation experiments on plywood.

\subsection{Dynamics of pushing and dragging}
Given the generalized coordinates of the object $\bm{q}=[x,y,\theta]^\mathsf{T}$, if we denote $m$ and $I = 1/6 m L^2$ to be the mass and moment of inertia of the object and denote $\bm{M} =$ diag($m,m,I$) to be its mass matrix, then the Newton-Euler equations for the object are: 
\begin{equation}
\bm{M \ddot{q} = F_p + F_f}
\label{eq:3}
\end{equation}
where $\bm{F_p} = [F_{x}, F_{y}, T_p]^\mathsf{T}$ is the force and torque applied by the pusher and $\bm{F_f} = [f_x, f_y, T_f]^\mathsf{T}$ is the frictional load applied by the surface under the object. The interaction force in normal direction between pusher and object is modeled with a penalty function method.

\subsection{Friction modelling}


We model the frictional forces  between the pusher, object and supporting surface.
%
%
If we denote $F_{\tau}$ and $F_n$ to be the tangential and normal forces between object and pusher, then the Coulomb friction law is:
\begin{equation}
-\mu_p  \le  F_{\tau}/F_n \le \mu_p
\label{eq:4}
\end{equation}

%
As shown in \figref{fig:AnisotropicFrictionModelling} (right), the contact patch between object and plywood is modeled as sets of rigidly connected point contacts. To be specific, we simulate the face contact as an $8\times8$ array of point contacts each subject to the anisotropic friction model in \eref{eq:1}.
The total frictional wrench $\bm{F_f} = [f_x, f_y, T_f]^\mathsf{T}$ is the sum of frictional wrenches from each point acting at the mass center of object. The normal load of each point is assumed to be $\frac{1}{64}$ of weight of the object.

\subsection{Parameter Identification}
\begin{figure}[t]
    \begin{minipage}[t]{0.53\linewidth}
    \centering
    \includegraphics[width=\linewidth]{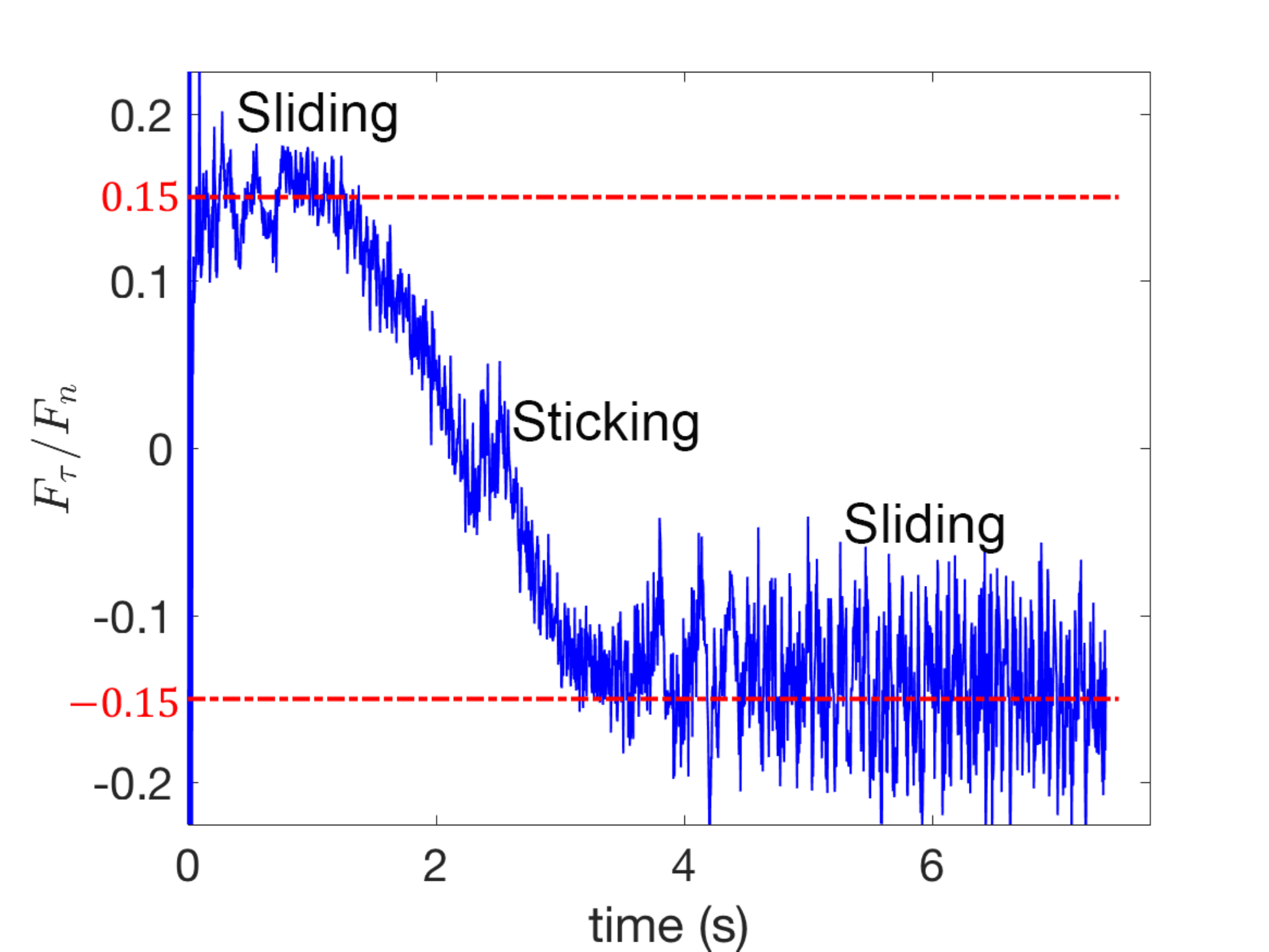}
    \label{fig: pusher friction_Coulomb}
    \end{minipage}%
    \begin{minipage}[t]{0.53\linewidth}
    \centering
    \includegraphics[width=\linewidth]{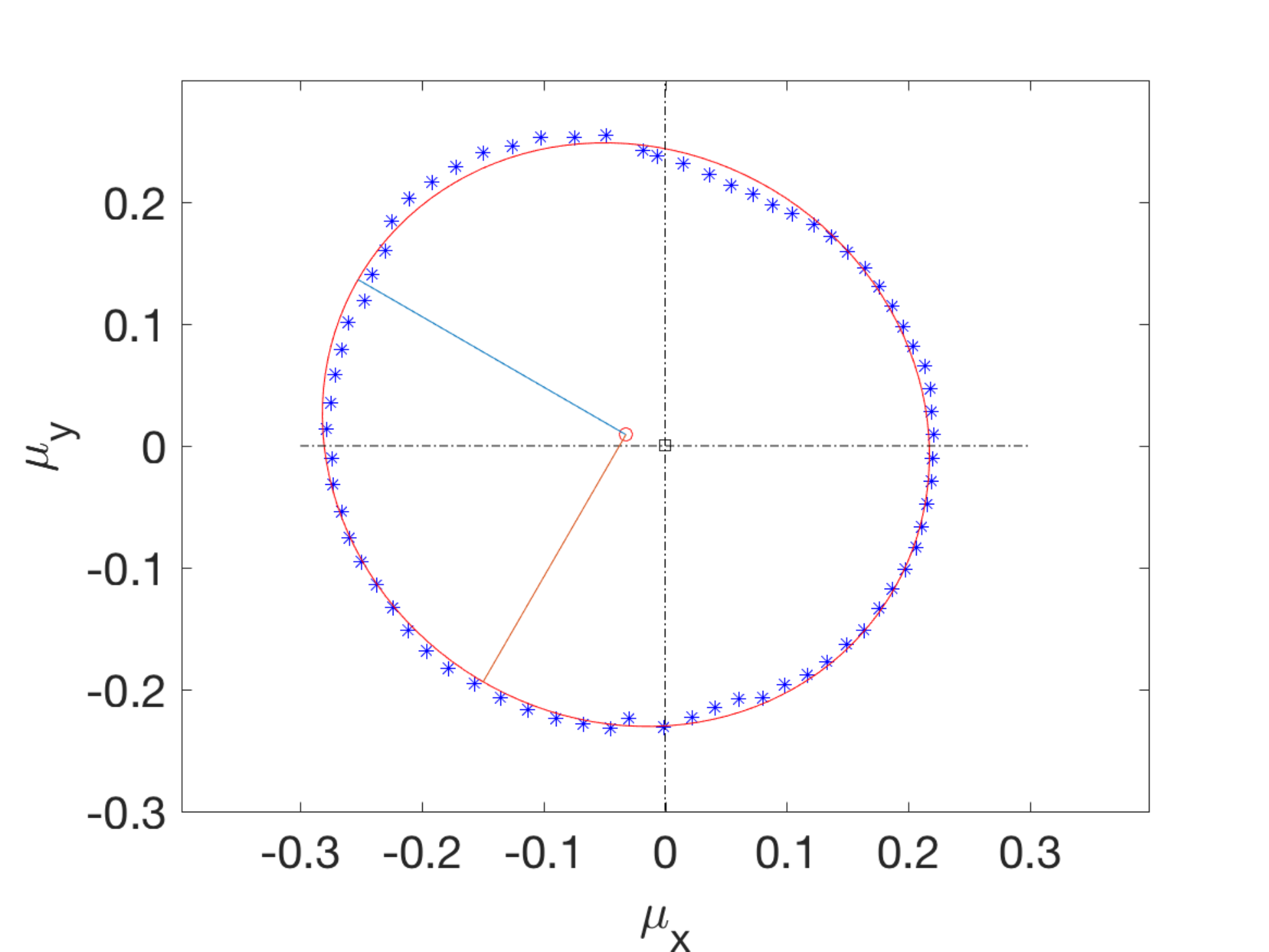}
    \label{fig: Fitted limit circle}
    \end{minipage}
    \caption{\change{(\textbf{Left}) Ratio of tangential and normal forces at the contact point between pusher and object indicates the contact follows Coulomb friction law. (\textbf{Right}) The fitted limit circle of anisotropic friction between object and plywood.}}
\end{figure}

To identify the friction parameters for the contact between pusher and object, we analyzed the ratio of tangential and normal forces at the pushing point for one pushing experiment. Fig. 10 (Left) shows that the ratio is $\pm 0.15$ when the contact point pair is sliding, while it lies between $-0.15$ and $0.15$ when the contact point pair is sticking. Thus Coulomb friction law can describe the friction between pusher and object approximately, with coefficient of friction $\mu_p \approx 0.15$.

\begin{table}[htbp]
    \centering
    \begin{tabular}{c|c|c|c|c}
        \hline
        \hline
        $\mu_a$ & $\mu_b$ & $m_0$ & $n_0$ & $\phi$ (rad)\\
        \hline
        \rowcolor[gray]{0.8} \begin{minipage}{1cm}\vspace{1mm} \centering  0.2545 \vspace{1mm} \end{minipage}  &  0.2346    &  0.0325    &  0.0082      & 2.6175\\
        \hline
    \end{tabular}    
    \caption{Parameters of anisotropic friction between steel and plywood (achieved by fitting with experiment data in Fig.6 of \cite{Yu2016MorePushing}).}
    \label{tab: fitted parameters}
\end{table}

For the contact between object and plywood surface, we identified the parameters of the elliptic limit surface via manual fitting, in \tabref{tab: fitted parameters}. Figure.10 (Right) compares the fitted limit ellipse with measured data.

\subsection{Simulation Results}
\label{sec:simulation result}
We carry out a simulation of the push-and-drag experiments in~\citep{Yu2016MorePushing} for 600  cycles in 6 batches, each consisting of 100 cycles with a different starting orientation: $0^{\circ}$, $60^{\circ}$, $120^{\circ}$, $180^{\circ}$, $240^{\circ}$ and $300^{\circ}$. 

Numerical results show that, independently of the initial orientation, the orientation of the pushing experiments converges to a stable direction in less than 50 cycles.

\begin{figure}[t]
    \begin{minipage}[t]{0.54\linewidth}
    \centering
    \includegraphics[width=\linewidth]{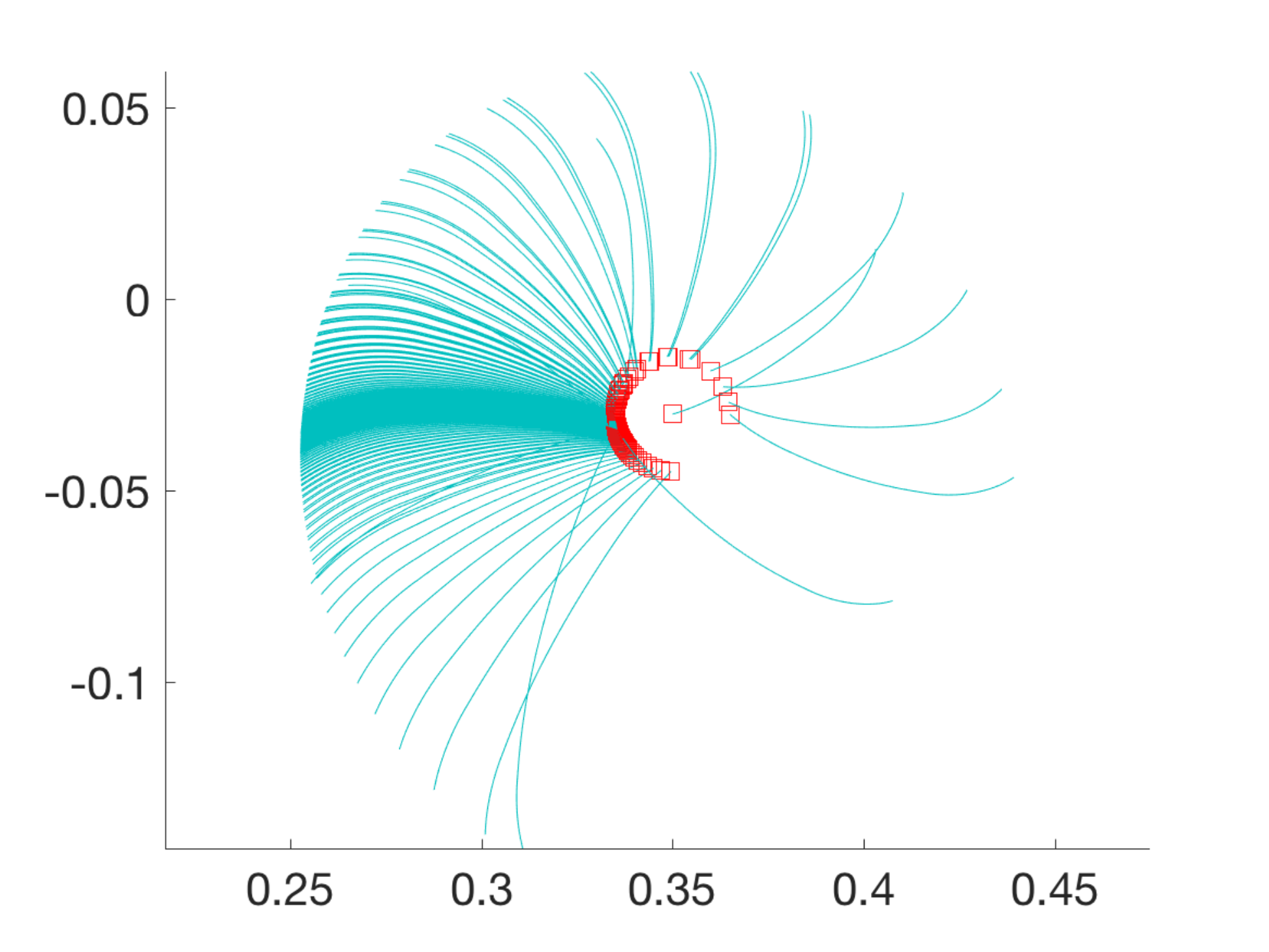}
    \end{minipage}%
    \begin{minipage}[t]{0.54\linewidth}
    \centering
    \includegraphics[width=\linewidth]{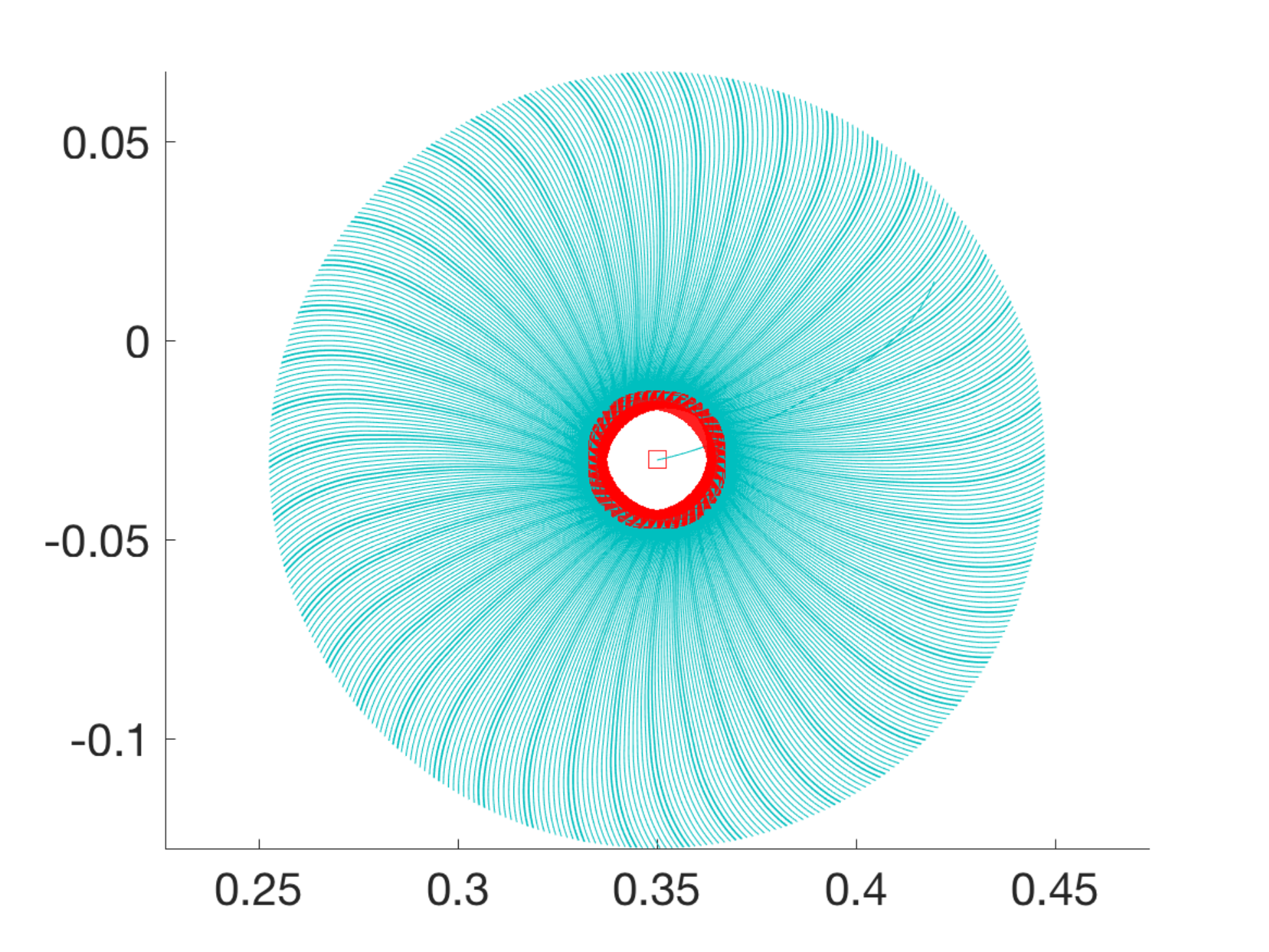}
    \end{minipage}
    \caption{Comparison of spatial distribution of trajectories of the center of mass of the pushed object in a global reference frame (in simulation) with an anisotropic (left) and isotropic (right) friction model between object and surface. The horizontal and vertical axis represent $x(m)$ and $y(m)$ respectively.\arnote{Is it possible to remove some of the labels in plots so that the plots are bigger? This is an important plot, but it is a bit small.}}
    \label{fig:CoM_ani_vs_iso}
\end{figure}

\begin{figure}[b]
    \begin{minipage}[t]{0.54\linewidth}
    \centering
    \includegraphics[width=\linewidth]{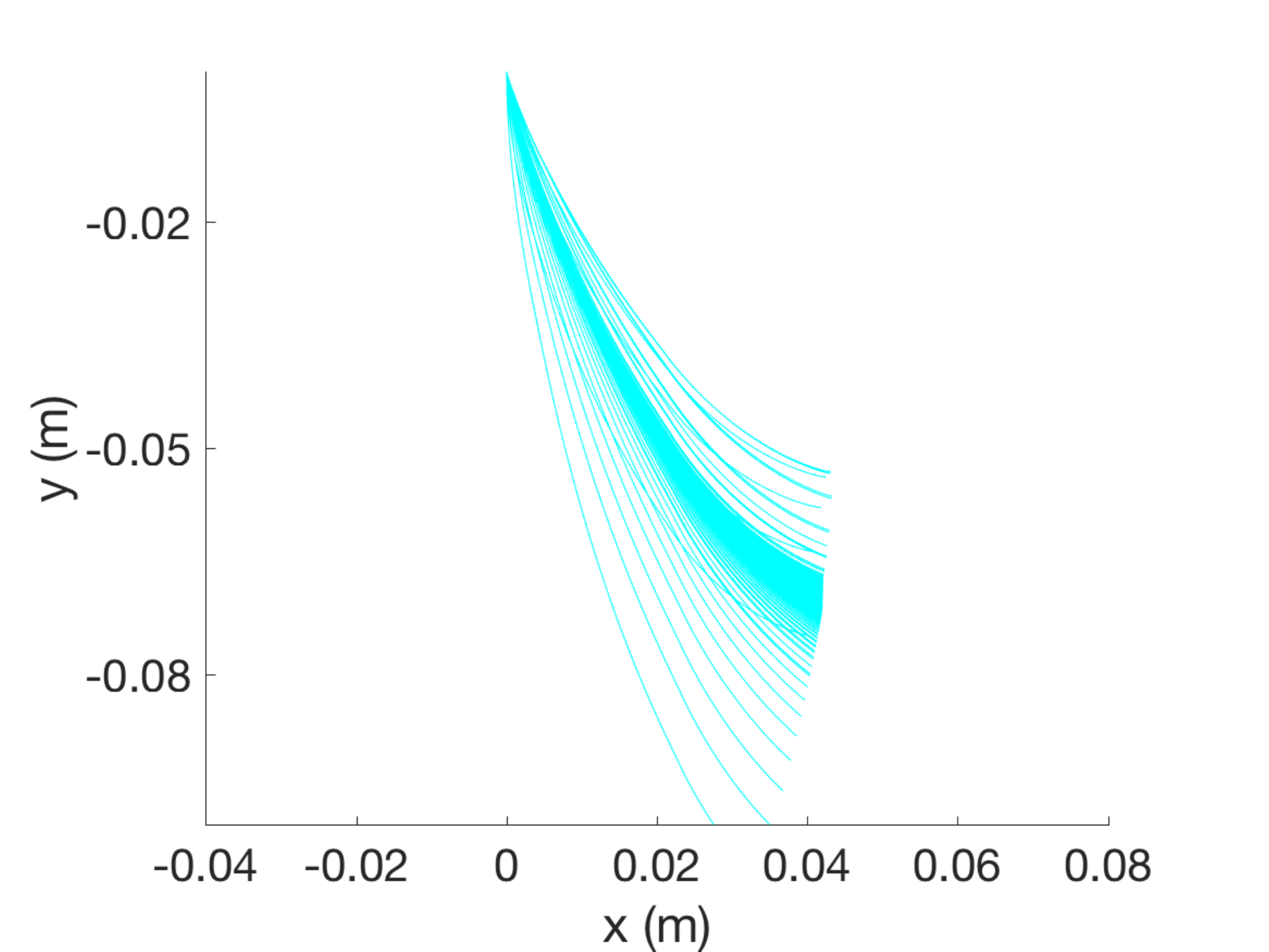}
    \end{minipage}%
    \begin{minipage}[t]{0.54\linewidth}
    \centering
    \includegraphics[width=\linewidth]{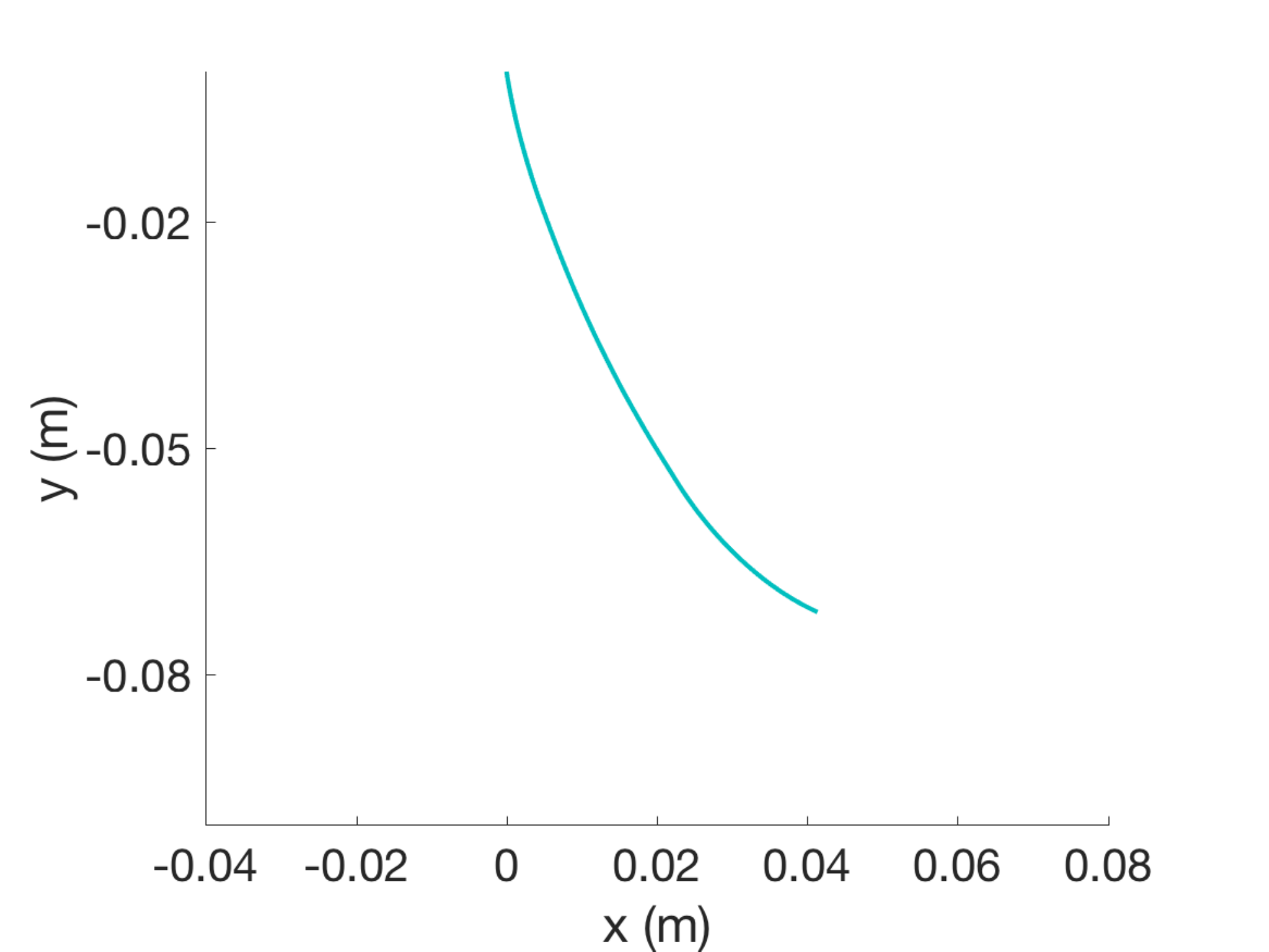}
    \end{minipage}
    \caption{(\change{ \textbf{Left})Simulated trajectories of the center of mass of a pushed object in the frame of the initial orientation of the object, subject to anisotropic friction. (\textbf{Right})The bold line is the equivalent outcome simulated with an isotropic friction model. All 500 trajectories coincide into a single one.}}
    \label{fig:CoM_ani_vs_iso_IOF}
\end{figure}

Figure~\ref{fig:CoM_ani_vs_iso} shows that, in contrast to an isotropic friction model, an anisotropic friction model generates significant biases in the distribution of initial orientations. 
Figure~\ref{fig:CoM_ani_vs_iso_IOF} shows the same trajectories but in the reference frame of the initial orientation of the object, recreating the plots in \figref{fig:Trajectories of CM in Object Base}.
This supports our hypothesis that anistotropic friction introduces sufficient non-linear dynamics in the data collection process to bias the collected dataset. This motivates the need for a more in depth analysis of the automated data collection process as a dynamics system itself.

\begin{figure}[t]
    \centering
    \includegraphics[width=0.8\linewidth]{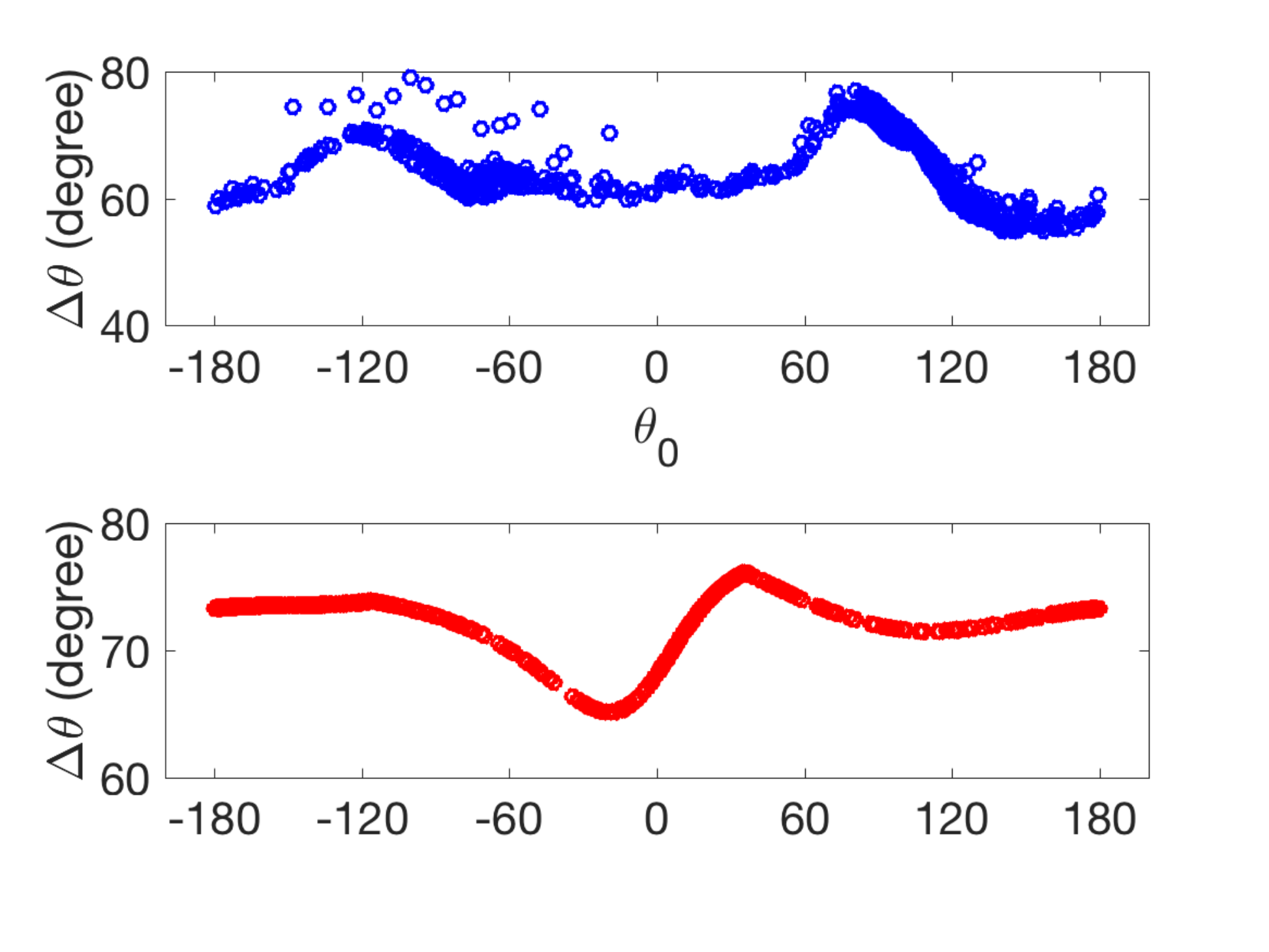}
    \caption{\change{Dependence of $\Delta\theta_0$ with the initial orientation $\theta_0$ of in simulation (red, bottom) for plywood compared with data in experiment (blue, top). The similarity trend indicates that anisotropic friction model can reproduces direction-dependent variance of pushed motion.}}
    \label{fig:Sim_vs_Exp_anisotropy}
\end{figure}

\change{We also simulate the pushed motion under an anisotropic friction law, and using the experimental distribution of initial orientations.
\figref{fig:Sim_vs_Exp_anisotropy} plots the dependence of $\Delta\theta$ with respect to the initial orientation $\theta_0$ which shows a similar structure when compared with experiment data on plywood. This indicates that anisotropic friction is key to explain the direction dependence of pushed motion.}

Finally, we would like to make a reference to recent work by \citet{Zhou2017AValidation}, where they model the stochasticity in pushed trajectories by sampling the coefficient of Coulomb friction from an interval, and yielding to a similar plot.
In this paper we suggest that the 'variance' in the coefficient of friction is likely due to a rather deterministic but anisotropic friction interaction.


\section{Conclusion and Discussion}

The main purpose of this paper is to understand the structured variability manifested in planar pushing dynamics~\citep{Yu2016MorePushing} and bring to light the importance of anisotropic friction. We focus the discussion in two aspects: the uncertainty in pushing experiments and the data collection dynamics.

\myparagraph{Anisotropic Friction and Structured Uncertainty.}
One of the key motivators for this paper is to better understand the nature of the uncertainty in the pushing trajectories in \figref{fig:Trajectories of CM in Object Base}.

We attribute it to three main factors: 1) a compressed representation of the state that projects contact dynamics to the initial reference frame of the object effectively lumping variability in the environment; 2) anisotropy of friction; and 3) sharp bias of the initial orientations of the collected data. 
The combination of these effects explains most of the structure in the noisy plots in \figref{fig:Trajectories of CM in Object Base}. \change{Remaining variability can be attributed to heterogeneity and aging of the surface.}

Standard deterministic models of friction dynamics with isotropic and uniform Coulomb friction imply a dynamic model of pushing that is invariant to the initial orientation of the object. But the detailed analysis indicates that anisotropy plays a key role in bending the pushing trajectories in different directions.
We simulate the pushing and dragging dynamics with an anisotropic friction law and show how bias and variability is formed. 

More efficient algorithms for identification of anisotropic friction parameters is an interesting topic we would like to exploit in the future. Although identification procedures for either parametric or non-parametric isotropic friction have been proposed, efficiency and accuracy are still open problems. 

\change{Micro-scale texture leads to asymmetries in friction. This could be exploited to generate useful directional behavior. We are interested in the problem of controlling anisotropic friction though embedding micro-textured patterns on contact surfaces. This opens the door to engineering friction for the purpose of robotic manipulation and locomotion.}

\myparagraph{Data collection dynamics}
Practical advances in machine learning and data-driven modeling are closely tied with big-data. The availability of large and nicely balanced data-sets is increasingly key to develop high performing systems.

Data collection with real robots in real scenarios, however, is much more challenging than in applications where simulation or computational data is sufficient.
We have seen that the dynamics of even a simple pushing automated data collection process can lead to significant biases in the dataset. 
%
%
As the system evolves, aging and wear is also a concern, which turns the dynamics that we are trying to capture a moving target.

In the case of this paper, the preference of certain initial conditions is a result of experiment design and anisotropy in the materials. 
Bias contributes to shape the variability of the dataset, which, if not dealt with, can result in deterioration of trained models. Hence the importance of paying attention to the data collection dynamics.

Particularly key for automating data collection in robotic manipulation is the availability of a resetting mechanism that can avoid bias.
Resetting a simulation experiment is trivial, but resetting the initial conditions of a real robotic task is more challenging.
Simple strategies as in~\citep{Yu2016MorePushing}, might leave the door open to bias. More complex resetting strategies that attempt to carefully control the initial conditions, might become as difficult to automate as the original problem we are trying to solve. Injecting controlled noise seems necessary. 

\change{Similar biases are present in other data collection experiments, which suggest the importance of a dynamic perspective on experimental data collection. 
We are interested in mathematical tools and mechanisms to track and control biases in data collection.}

\bibliographystyle{IEEEtranN} 
{\footnotesize\bibliography{dm-icra18}}
\end{document}